\documentclass{article}
\usepackage{amsmath,amsfonts,bm}

\def\eqref#1{equation~\ref{#1}}
\def\1{\bm{1}}

\DeclareMathAlphabet{\mathsfit}{\encodingdefault}{\sfdefault}{m}{sl}
\SetMathAlphabet{\mathsfit}{bold}{\encodingdefault}{\sfdefault}{bx}{n}

\PassOptionsToPackage{numbers}{natbib}
\usepackage[preprint]{neurips_2026}
\usepackage{times}
\usepackage{hyperref}
\usepackage{graphicx}
\usepackage{booktabs}
\usepackage{multirow}
\usepackage{longtable}
\usepackage{threeparttable}
\usepackage{adjustbox}
\usepackage{url}
\usepackage{float}
\usepackage{enumitem}
\usepackage{subcaption}
\usepackage{graphicx}
\usepackage[dvipsnames]{xcolor}
\usepackage{cleveref}
\crefname{equation}{eqn.}{eqns.}

\usepackage{titlesec}
\titlespacing*{\section}{0pt}{1.2ex plus .25ex minus .2ex}{0.8ex plus .15ex}
\titlespacing*{\subsection}{0pt}{0.9ex plus .2ex minus .15ex}{0.5ex plus .12ex}
\titlespacing*{\subsubsection}{0pt}{0.8ex plus .18ex minus .12ex}{0.35ex plus .08ex}
\titlespacing*{\paragraph}{0pt}{0.6ex plus .15ex minus .08ex}{0.6em}

\AtBeginDocument{%
  \setlength{\abovedisplayskip}{3pt}%
  \setlength{\belowdisplayskip}{3pt}%
  \setlength{\abovedisplayshortskip}{2pt}%
  \setlength{\belowdisplayshortskip}{2pt}%
  \setlength{\jot}{2pt}% reduce inter-line gap in aligned equations
  \setlength{\textfloatsep}{8pt}%
  \setlength{\floatsep}{6pt}%
  \setlength{\intextsep}{8pt}%
  \setlength{\abovecaptionskip}{4pt}%
  \setlength{\belowcaptionskip}{4pt}%
}

\title{Tabular Numeric Stretch Transformation}

\author{
Zihao Ye \quad
Juyong Kim \quad
Johnna Sundberg \quad
Burak Varici \quad
Pradeep Ravikumar
\\
Machine Learning Department \\
Carnegie Mellon University \\
Pittsburgh, PA 15213, USA \\
\texttt{
\{zihaoye,juyongk,jsundber,bvarici\}@andrew.cmu.edu
} \\
\texttt{pradeepr@cs.cmu.edu}
}

\newcommand{\update}[1]{#1}
\newcommand{\n}[1]{#1}
\newcommand{\BV}[1]{#1}

\begin{document}

\maketitle

\begin{abstract}
\n{Tabular data presents unique challenges for deep learning due to its heterogeneous nature, where numeric features exhibit diverse distributions, scales, and statistical properties. Although recent advances have improved how models learn from tabular data, how numeric data are transformed into model-friendly representations remains comparatively underexplored.}
\n{We introduce the \emph{stretch transformation framework}, which formulates numeric feature preprocessing as an optimization problem to make the target function smoother and thus more learnable.} Our framework has two variants: (1) \emph{unsupervised stretch}, which uniformly redistributes feature density via minimax optimization, and (2) \emph{supervised stretch}, \n{which optimizes target-aware numeric feature transformations from the perspective of target-function smoothness by minimizing the target function's Dirichlet energy in the transformed space.}
\n{Our theoretical analysis further connects this framework to several popular transformations: unsupervised stretch is closely related to Piecewise Linear Encoding through a shared piecewise-linear geometry and approaches the empirical CDF transformation as the number of bins grows, while supervised stretch becomes closely related to target encoding in the fine-binning limit.}
Comprehensive experiments on 38 datasets from the TALENT benchmark demonstrate that supervised stretch consistently outperforms all baselines. These results show that explicitly optimizing for target function smoothness is a powerful and underexplored strategy for tabular deep learning.
\end{abstract}

\section{Introduction}

\n{Most recent progress in tabular deep learning has focused on improving how models learn from data through better architectures, optimization, or training protocols. A complementary question is how to transform numeric data itself into representations that are easier for neural networks to learn from. This question is important because numeric tabular features can combine heterogeneous quantities, binary flags, counts, percentages, monetary values, and heavy-tailed continuous variables, each with distinct scales and statistical properties \citep{whydotree}.} This heterogeneity creates a challenging optimization landscape for neural networks, which typically assume well-conditioned inputs \citep{batchnorm}.

While tree-based methods naturally handle diverse distributions through split-based decisions, neural networks require careful preprocessing to achieve competitive performance \citep{talent1,talent2}. Thus, \n{the \emph{numeric feature transformation} step} is needed to bridge the gap between raw heterogeneous data and model-friendly features. Commonly used transformations include standardization, power transformations~\citep{yeo_johnson}, quantile transformations~\citep{scikit}, and Piecewise Linear Encoding (PLE)~\citep{ple}. Remarkably, although target information (e.g., labels) is typically available during training, \n{most existing numeric feature transformations are unsupervised. A supervised PLE variant, PLE-T, is a notable exception, but it uses targets only to set bin boundaries, not to optimize the transformation itself from the perspective of target-function learnability.}

We introduce the \emph{stretch transformation framework}, which systematically incorporates target information to create more learnable \n{numeric} representations. Our key insight is that feature transformation should not merely normalize distributions but make the target function smoother in the transformed space. 
\BV{This reduces the design of numeric feature transformations to a width-allocation problem: how much transformed-space width should be assigned to different regions of a numeric feature? We answer this question via two complementary approaches:}
%\n{The framework asks a single design question: how should transformed-space width be allocated across numeric feature regions? We answer this question differently depending on whether target information is used.} We present two complementary approaches:

\noindent\textbf{Supervised stretch} is, to our knowledge, \n{the first method to formulate target-informed numeric feature transformation explicitly as target-function smoothness optimization.} By minimizing the Dirichlet energy of the target function in the transformed space, we derive optimal width allocations that concentrate more resolution in regions where the target varies rapidly. This principled approach achieves strong empirical gains, demonstrating the untapped potential of supervised feature transformation. \looseness=-1

\noindent\textbf{Unsupervised stretch} provides a robust fallback when target information is unavailable or unreliable. Using minimax optimization, it maximizes the worst-case sample separation, effectively creating a piecewise linear approximation of the empirical CDF transformation. This variant matches or exceeds PLE's performance while requiring only $O(1)$ memory per feature instead of $O(T)$.

Beyond practical improvements, our framework offers theoretical explanations for existing empirical observations. We show that unsupervised stretch explains why CDF transformation can reduce frequency content and improve learnability~\citep{tabularfreq}, despite being label-agnostic. Similarly, supervised stretch reveals deep connections to target encoding, providing a theoretical justification for why target-based transformations can be effective while offering principled regularization through the number of bins $T$. \textbf{In summary, our contributions are:}
\begin{itemize}[leftmargin=1.5em, topsep=0pt, itemsep=0pt]
    \item We introduce supervised stretch, a systematic framework that uses target information to optimize numeric feature transformations by maximizing target-function smoothness.
    \item We propose unsupervised stretch, a memory-efficient alternative to PLE that achieves comparable or superior performance without dimensional expansion.
    \item \n{We connect our framework to several existing transformations: unsupervised stretch shares a piecewise-linear geometry with PLE and approaches the empirical CDF transformation as the number of bins grows, and supervised stretch becomes closely related to target encoding in the fine-binning limit.}
    \item \n{We empirically validate the framework on 38 datasets from the TALENT benchmark, where supervised stretch consistently outperforms all baselines, with the largest gains in regression tasks.}
    % \item The first systematic framework for incorporating target information into 
    % numeric feature transformation through smoothness optimization.
    % \item Theoretical analysis connecting our methods to CDF transformation and target 
    % encoding, explaining their empirical effectiveness.
    % \item Comprehensive experiments on 38 datasets showing supervised stretch consistently 
    % outperforms all baselines, with particularly strong gains in regression tasks.
    % \item A memory-efficient alternative to PLE that achieves comparable performance 
    % without dimensional expansion.
\end{itemize}

These results challenge the prevailing paradigm of unsupervised feature transformation in tabular deep learning and open new avenues for supervised preprocessing methods.

\section{Related Work}\label{sec:related}

\noindent\textbf{Tabular Machine Learning} is a rapidly evolving field \citep{vanbreugel2024tabularfoundationmodelsresearch, Borisov_2024, shwartzziv2021tabulardatadeeplearning, hancock}. Recent work has proposed various neural architectures for tabular data, including attention-based models \citep{ftt, tabnet} and MLP-based models \citep{ple, realmlp, tabm}, achieving strong performance on standard benchmarks \citep{erickson2025tabarenalivingbenchmarkmachine}. Beyond architectural innovations, several studies have explored pretraining for large-scale tabular datasets \citep{carte}, while recent advances in in-context learning have enabled foundation models to perform well on small datasets without task-specific finetuning \citep{tabpfn, tabpfnv2, tabicl}.

%\noindent\textbf{Feature Transformation for Tabular Data.} Effective feature preprocessing is crucial for tabular machine learning, as heterogeneous features often exhibit disparate scales and distributions \citep{whydotree}. As a formation of multimodal data, tabular features encompass binary indicators, count data, and heavy tailed continuous distributions, necessitating tailored transformation strategies. 
\noindent\textbf{Feature Transformations for Tabular Data.} Effective preprocessing is crucial for tabular machine learning, as heterogeneous features, such as binary indicators, counts, and heavy-tailed continuous variables with disparate scales, pose distinct challenges for neural networks~\citep{whydotree}. This complexity necessitates feature transformation strategies tailored to tabular data. While tree-based models (e.g., Random Forests, Gradient Boosted Trees) are invariant to monotonic transformations~\citep{xgboost}, neural networks often require standardized inputs to stabilize optimization~\citep{batchnorm}.

Standardization (z-score normalization) remains the most common preprocessing technique, mapping each feature to zero mean and unit variance \citep{scikit}. %However, it only rescales and shifts distributions without addressing skewness or outliers. 
but assumes approximate normality and offers no robustness to skewness or outliers.
Power transformations address non-normality: the Yeo-Johnson transformation \citep{yeo_johnson} extends Box-Cox \citep{boxcox} to handle negative values, producing more Gaussian-like features, which benefits neural networks and has been widely adopted in tabular machine learning \citep{tabpfn}. Piecewise linear encoding (PLE) \citep{ple} discretizes the distribution into quantile-based bins and encodes each into a vector, which has also proven effective in tabular tasks.

\noindent\textbf{Neural Network Inductive Biases.} Despite being universal approximators \citep{uat1, uat2}, neural networks exhibit specific inductive biases that can influence their performance in tabular data. For instance, the spectral bias phenomenon, whereby ReLU networks preferentially learn low-frequency functions before high-frequency components, has been well-documented \citep{spectralbias, f-principle}. Interestingly, \cite{tabularfreq} observe that tabular data contains substantially higher frequency components compared to natural images. \update{Unlike images defined on a fixed pixel grid, numeric features imply continuity, allowing samples to be arbitrarily close to each other. This characteristic creates unbounded high-frequency components that are notoriously difficult for neural networks to learn due to spectral bias.} \n{This observation provides one lens through which to understand the empirical performance gap between neural networks and tree-based methods on tabular benchmarks, and motivates preprocessing methods that change not only feature marginal distributions, but also the smoothness of the induced feature-target map.}

\section{Stretch Feature Transformation}\label{sec:stretch-transform}

\begin{figure}[t]
    \begin{center}
    \includegraphics[width=0.95\linewidth]{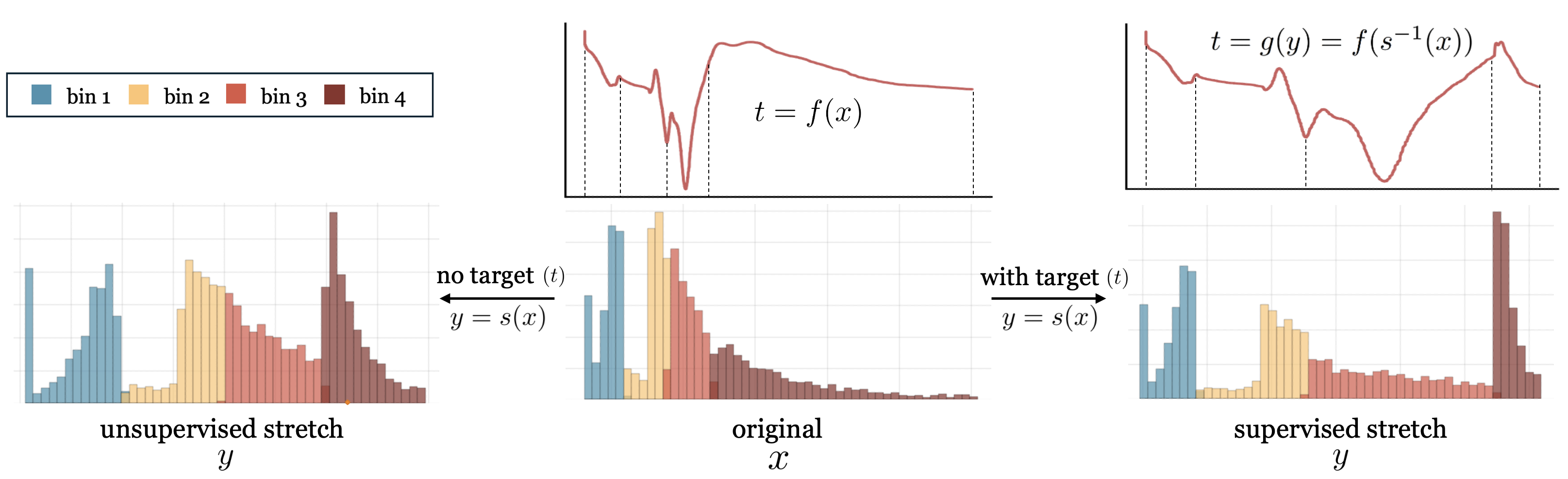}
    \caption{Overview: Stretch Transformation Framework. \textbf{(Center)} An original feature $x$ with a complex target function $f(x)$, partitioned into four quantile bins (colors). \textbf{(Left)} Unsupervised stretch uniformly redistributes feature density. \textbf{(Right)} Supervised stretch uses target information to allocate more \emph{stretch} to bins where the target function varies rapidly (e.g., bin 3). This creates a new target function $g(y)$ in the transformed space that is significantly smoother and thus easier for a neural network to learn.}
    \end{center}
    \label{fig:overview}
\end{figure}
At a high level, stretch transformations preserve the ordering of each numeric feature while reallocating transformed-space length across regions. They can expand regions where the target changes rapidly and compress regions with little target-relevant variation.
%At a high level, stretch transformations keep the ordering of each numeric feature fixed but choose how much transformed-space length each region receives. This lets the transformation expand regions where the target changes rapidly and compress regions where little target-relevant variation is present.

Denote a tabular dataset by $\mathcal{D} = \{(\mathbf{x}_i, t_i)\}_{i=1}^n$, where $\mathbf{x}_i \in \mathbb{R}^d$ are features and $t_i$ is the corresponding target.\footnote{\label{fn:target}For simplicity, we write $t_i$ as the target. Our framework is directly applicable to both scalar targets (e.g., class labels, single value regression) and multidimensional vector targets ($\mathbf{t}_i \in \mathbb{R}^k$).} While targets are available during training, feature transformations may leverage this information (supervised) or ignore it (unsupervised), though most existing methods are unsupervised. A feature transformation $s: \mathcal{X} \rightarrow \mathcal{Y}$ preprocesses the raw features, creating the learning pipeline: \looseness=-1
\begin{equation}\label{eq:prelim}
    \mathbf{x} \xrightarrow{s} \mathbf{y} = s(\mathbf{x}) \xrightarrow{g_\theta} \hat{t} = g_\theta(\mathbf{y})
\end{equation}
For an invertible transformation with the inverse $h = s^{-1}$, the target function in the transformed space becomes $g(y) = \mathbb{E}[t|y] = f(h(y))$ where $f(x) = \mathbb{E}[t|x]$ is the original target function. The neural network $g_\theta$ learns to approximate $g$ in the transformed space. The goal is to find $s$ such that $g = f \circ h$ is easier for neural networks to learn than the original $f$.

%Feature transformations vary along multiple axes. Unsupervised transformations (e.g., standardization, quantile transformation) use only feature statistics, while supervised transformations leverage target information during design; they may be dimension-preserving ($d' = d$) or dimension-expanding ($d' > d$) like PLE; and marginal transformations operate on features independently, while joint transformations model feature interactions.
Feature transformations can be characterized along several axes: they may be unsupervised (using only feature statistics, e.g., standardization, quantile transformation) or supervised (leveraging target information); dimension-preserving ($d' = d$) or dimension-expanding ($d' > d$, as in PLE); and marginal, operating on features independently, or joint, modeling feature interactions. These properties can be combined as well, for instance, distribution-shaping followed by standardization.

For numeric features, desirable properties of a transformation $s$ include: (1) \emph{monotonicity}, preserving the natural ordering to avoid arbitrary permutations that could destroy meaningful relationships; (2) \emph{bounded output range}, e.g., $[0,1]$ or $[-1,1]$, to ensure inputs are well-scaled for neural networks.

Our stretch transformation framework, introduced shortly, provides both unsupervised and supervised variants that satisfy these requirements through a piecewise linear design. The framework operates marginally on each feature while preserving dimensionality. For the supervised variant, target information is used \emph{only} during transformation design; the learned transformation is then fixed and applied identically to all data.

\subsection{Stretch Transformation Setup}\label{sec:framework}

We introduce a feature transformation that rescales different regions of the input via a monotone piecewise linear map. For a scalar feature $x\in\mathbb{R}$ with range $[x_{\min},x_{\max}]$, we first partition the domain into $T$ intervals with boundaries $\mathbf{b}=\{b_0,b_1,\ldots,b_T\}$ where $b_0=x_{\min}$ and $b_T=x_{\max}$. We adopt quantile-based binning as the default partition, ensuring approximately equal sample counts per bin. This process groups samples $\{x_i\}_{i=1}^n$ (assumed to be sorted) into $T$ consecutive bins. We denote the set of sample indices $i$ whose corresponding feature value $x_i$ falls into the $t$-th bin, $[b_{t-1}, b_t)$, as $\mathcal{I}_t$. The transformation $s:\mathbb{R}\to[0,1]$ is then defined by
\begin{equation}
    s(x)=c_{t-1}+\frac{x-b_{t-1}}{b_t-b_{t-1}}\cdot w_t,\qquad x\in[b_{t-1},b_t) \ ,
\end{equation}
where $w_t>0$ is the width allocated to interval $t$, satisfying $\sum_{t=1}^T w_t=1$. The term $c_t:=\sum_{i=1}^t w_i$ (with $c_0=0$) denotes the cumulative width. This form is strictly increasing (hence order-preserving and bijective onto $[0,1]$). The slope within each interval is
\begin{equation}
    s'(x) = \frac{w_t}{b_t-b_{t-1}} \ ,  \quad  \forall x\in[b_{t-1},b_t), \; t \in [T] \ .
\end{equation}
Thus, by choosing $\{w_t\}$, we can control the local stretch (large $w_t$) or squeeze (small $w_t$) of each region. \textbf{The central design problem is how to set the width vector $\{w_t\}_{t=1}^T$, which we derive from a principled optimization objective.}

\subsection{\n{Optimization Objective: Dirichlet Energy Minimization}}\label{sec:objective}

\n{Recall from \Cref{eq:prelim} that $g(y) = f(h(y))$ is the target function in the transformed space, where $f$ is vector-valued for classification.} A smoother $g$ in the $y$-space is easier for neural networks to learn due to their spectral bias toward low-frequency functions~\citep{f-principle,spectralbias}. Thus, we seek transformations that maximize the smoothness of~$g$. \n{To make this smoothness objective precise, we adopt a kernel-smoothing perspective: a function is smoother if it changes less under small-bandwidth smoothing.} We derive this connection below, with the full derivation provided in~\Cref{app:dirichlet-motivation}. To quantify smoothness, we consider the kernel correlation with a Gaussian RBF kernel $k_\sigma(y,z) = \exp(-(y-z)^2/(2\sigma^2))$: \looseness=-1
\begin{equation}\label{eq:kernel-corr-with-rbf}
    \langle g, T_\sigma g\rangle = \int_0^1\int_0^1 g(y)k_\sigma(y,z)g(z)\,dy\,dz \ ,
\end{equation}
where $T_\sigma$ is the integral operator associated with $k_\sigma$. In the small bandwidth regime ($\sigma \to 0$), the heat kernel expansion \citep{varadhan1967behavior,molchanov1975diffusion} gives $T_\sigma = e^{\frac{\sigma^2}{2}\partial_{yy}} = I + \frac{\sigma^2}{2}\partial_{yy} + O(\sigma^4)$. To isolate the term independent of smoothness, we use the centered version:
\begin{equation}\label{eq:kernel-corr-with-rbf-centered}
    \langle g, (T_\sigma - I)g\rangle = \langle g, \frac{\sigma^2}{2}\partial_{yy}g\rangle + O(\sigma^4) = -\frac{\sigma^2}{2}\int_0^1 \|g'(y)\|_2^2\,dy + O(\sigma^4),
\end{equation}
where the last equality follows from integration by parts. Therefore, maximizing kernel smoothness is equivalent to minimizing the Dirichlet energy:
\begin{equation}\label{eq:minimize-dirichlet}
    \min_{h}\ \mathcal{E}[g] := \int_0^1 \|g'(y)\|_2^2\,dy \ .
\end{equation}
Recall that our stretch transformation allocates width $w_t$ to bin $t$, controlling how much of the transformed space $[0,1]$ is dedicated to each region. Given samples $\{x_i\}_{i=1}^n$ with transformed positions $y_i = s(x_i)$, using the piecewise-linear interpolant through the transformed samples gives the following finite-sample discrete Dirichlet energy:
\begin{equation}\label{eq:disc-dirichlet}
    \mathcal{E}_{\mathrm{disc}} = \sum_{i=1}^{n-1}\frac{\|\Delta f_i\|_2^2}{\Delta y_i}, \quad \text{where} \quad \Delta f_i := f(x_{i+1}) - f(x_i), \quad \Delta y_i := y_{i+1} - y_i.
\end{equation}
This objective depends on both the target increments $\Delta f_i$ and the \n{spacings} in the transformed space~$\Delta y_i$. The spacings $\Delta y_i$ are determined by our width allocations: within bin $t$ with width $w_t$, the spacings sum to $w_t$. The key question then is how to choose $\{w_t\}$ to minimize this energy, which we answer in two distinct settings of unsupervised and supervised.

\subsection{Unsupervised Stretch: Uniform Allocation}\label{sec:unsupervised}

Without the target information, we cannot directly evaluate $\Delta f_i$ in~\Cref{eq:disc-dirichlet}. Instead, we adopt a minimax principle and minimize the worst-case Dirichlet energy over all target functions with bounded local variation. Specifically, assuming $\|\Delta f_i\|_2^2 \leq C$ for all $i$, the worst-case Dirichlet energy for spacing allocation $\{\Delta y_i\}$ becomes
\begin{equation}
    \max_{\{\Delta f_i\}} \sum_{i=1}^{n-1}\frac{\|\Delta f_i\|_2^2}{\Delta y_i} \quad \text{s.t.} \quad \|\Delta f_i\|_2^2 \leq C, \ \forall i.
\end{equation}
The maximizer is straightforwardly $\|\Delta f_i\|_2^2 = C$ for all $i$, leading to the optimization problem (see Appendix~\ref{app:unsupervised-objective-derivation} for detailed derivation):
\begin{equation}\label{eq:worst-case-dirichlet-unsupervised}
    \min_{\{\Delta y_i\}} \sum_{i=1}^{n-1} \frac{1}{\Delta y_i} \quad \text{s.t.} \quad \sum_{i=1}^{n-1} \Delta y_i = 1 \ .
\end{equation}
Within our binning framework with $T$ quantile bins containing $n_t \approx n/T$ samples each, the optimal allocation is then uniform both within and across bins, yielding:
\begin{equation}\label{eq:optimal-uniform-allocation}
    w_t^\star = \frac{1}{T} \ , \qquad \forall t\in\{1,\ldots,T\} \ .
\end{equation}
\noindent\textbf{Connection to empirical CDF.} With this uniform allocation, unsupervised stretch becomes a piecewise linear approximation of the empirical CDF transformation. As $T \to n$, our method converges to the exact empirical CDF, which maps each sample to its normalized rank. This insight provides a theoretical justification for the empirical finding in \citet{tabularfreq} that CDF transformation reduces frequency content and improves learnability. By maximizing the minimum spacing between samples, we effectively smooth the target function, reducing the high-frequency components.
However, as noted in \citet{tabularfreq}, reducing high-frequency components does not guarantee a better performance, since excessive smoothing can eliminate important signals. The parameter $T$ provides a natural trade-off: a small $T$ preserves more of the original distribution's structure, while a large $T$ approaches full CDF transformation. This controllable interpolation between preserving distributional features and achieving uniform density is a key advantage of our framework.

\noindent\textbf{Connection to Piecewise Linear Encoding (PLE).} Although unsupervised stretch and PLE \citep{ple} appear fundamentally different—scalar versus $T$-dimensional outputs—they share an underlying geometric structure. PLE maps a sample $x$ in bin $t$ to:
\begin{equation}
    \text{PLE}(x) = [1, \ldots, 1, \frac{x - b_{t-1}}{b_t - b_{t-1}}, 0, \ldots, 0] \in \mathbb{R}^T \ .
\end{equation}
This creates a piecewise linear path in $\mathbb{R}^T$ from origin to $(1,\ldots,1)$. The arc length along this path is \looseness=-1
\begin{equation}
    L(x) = (t-1) + \frac{x - b_{t-1}}{b_t - b_{t-1}} = T \cdot \text{Unsupervised-Stretch}(x) \ .
\end{equation}
Thus, unsupervised stretch precisely parameterizes the PLE manifold by normalized arc length. Both achieve identical quantile-based density redistribution, differing only by coordinate representation. This equivalence explains their similar empirical performance (Section~\ref{sec:analysis}), while unsupervised stretch offers significant computational advantages: $O(1)$ versus $O(T)$ memory per feature and no dimensional expansion.

\subsection{Supervised Stretch: Target-Informed Allocation}
\label{sec:supervised}

When target information is available, we can optimize the discrete Dirichlet energy in~\Cref{eq:disc-dirichlet} directly. However, substituting the piecewise linear map proves to be numerically unstable, as the resulting objective is highly sensitive to the original feature spacings $\Delta x_i$ (see Appendix~\ref{app:supervised-objective-derivation} for a detailed discussion). To overcome this, we instead optimize a robust lower bound on the energy.

We partition the samples into $T$ bins. For bin $t$ with width $w_t = \sum_{i \in \mathcal{I}_t}\Delta y_i$, let $S_t := \sum_{i \in \mathcal{I}_t}\|\Delta f_i\|_2$ be the bin-wise total variation of the target. For fixed $w_t > 0$, minimizing the within-bin contribution of~\Cref{eq:disc-dirichlet} over $\{\Delta y_i\}_{i \in \mathcal{I}_t}$ subject to $\sum_{i \in \mathcal{I}_t}\Delta y_i = w_t$ yields a lower bound:\looseness=-1
\begin{equation}
    \sum_{i \in \mathcal{I}_t}\frac{\|\Delta f_i\|_2^2}{\Delta y_i} \geq \frac{S_t^2}{w_t} \ ,
\end{equation}
with equality at $\Delta y_i \propto \|\Delta f_i\|_2$ within each bin. Then, the global optimization over $\{w_t\}$ becomes
\begin{equation}\label{eq:supervised-optimization}
    \min_{\{w_t > 0\}} \sum_{t=1}^T \frac{S_t^2}{w_t} \quad \text{s.t.} \quad \sum_{t=1}^T w_t = 1 \ .
\end{equation}
This convex problem has the closed-form solution:
\begin{equation}\label{eq:l2-widths}
    w_t^\star = \frac{S_t}{\sum_{u=1}^T S_u}.
\end{equation}
Intuitively, in the transformed space, we allocate a larger space to the regions where the target function varies rapidly, effectively equalizing the slope magnitudes across bins and creating a smoother target function in the transformed space.

\noindent\textbf{Connection to target encoding.} Supervised stretch connects naturally to target encoding when $T = n$ (one bin per unique value). In this limit, the optimal width for the interval $[x_i, x_{i+1}]$ becomes:
\begin{equation}
    w_i = \frac{|f_{i+1} - f_i|}{\sum_{k=1}^{n-1}|f_{k+1} - f_k|} \ .
\end{equation}
This is remarkably similar to applying min-max scaling to target encoding. Standard target encoding maps $x_i \mapsto f_i = \mathbb{E}[t|x=x_i]$, and if we scale these values to $[0,1]$, we get essentially the same transformation for monotonic targets.
The key insight is that both methods fundamentally use target variation to guide the transformation, target encoding does it directly, while supervised stretch achieves it via Dirichlet energy minimization. 
Our framework thus provides a theoretical justification for why target-based transformations are effective: they implicitly smooth the target function in the transformed space.

While target encoding is widely applied to categorical features, its use for numeric features has been limited, partly due to concerns about overfitting or interpretability. The proposed supervised stretch offers a regularized, theoretically grounded approach to incorporating target information for numeric features, with the number of bins $T$ serving as a natural regularization parameter.

\noindent\textbf{Out-of-fold estimation.} In practice, to prevent information leakage, we use $K$-fold cross-validation with adaptive Nadaraya-Watson kernel regression to obtain $\widehat{f}(x_i)$ for each sample without using its own target value \citep{KernelRegression, KernelRegression2} (details in Appendix~\ref{app:oof-implementation}). The bin widths are computed using $\Delta\widehat{f}_i$ in place of $\Delta f_i$.

\section{Experiments}\label{sec:experiments}

We evaluate the proposed stretch transformations against established baselines across comprehensive tabular benchmarks, assessing their impact on diverse neural architectures. \n{The experiments are designed to answer three questions: whether target-aware smoothness optimization improves numeric feature transformation, whether the gains depend on task type, and how robust the conclusions are across models and evaluation criteria.} \n{Our results show that supervised stretch has the strongest aggregate performance, with particularly clear advantages in regression.} Detailed experimental configurations and additional results are provided in Appendix~\ref{app:experiment-setup}.

\subsection{Experimental Setup}

\noindent\textbf{Datasets and Models.}
We use the TALENT benchmark suite \citep{talent1, talent2},  focusing on the \emph{Tiny Benchmark 1} collection of 26 classification and 12 regression datasets. This benchmark covers diverse tabular learning tasks with varying dataset sizes, feature types, and target distributions. Categorical features are encoded as indices by default, with RealMLP internally converting them to one-hot encoding as per its design \citep{realmlp}. Our investigation focuses on numeric feature transformations. \update{Detailed characteristics of these datasets, including the breakdown of numeric versus categorical feature proportions, sample sizes, and feature counts, are provided in Appendix~\ref{app:datasetcomposition}.}

We evaluate five representative neural architectures for tabular data: FT-Transformer (FTT)~\citep{ftt}, standard MLP, MLP-PLR~\citep{ple}, RealMLP~\citep{realmlp}, and ResNet~\citep{resnet}. Experiments span 38 datasets, yielding 190 dataset-model combinations. This is a controlled transformation comparison rather than an end-to-end model leaderboard. For RealMLP, the RS-SC cell retains its original \texttt{RobustScale+SmoothClip} numeric preprocessing. For every alternative transformation, we disable the internal RS-SC stage so that the transformations are not stacked; model and training hyperparameters are nevertheless tuned under the same Optuna protocol in every cell.

\begin{figure}[t]
    \begin{center}
    \includegraphics[width=\linewidth]{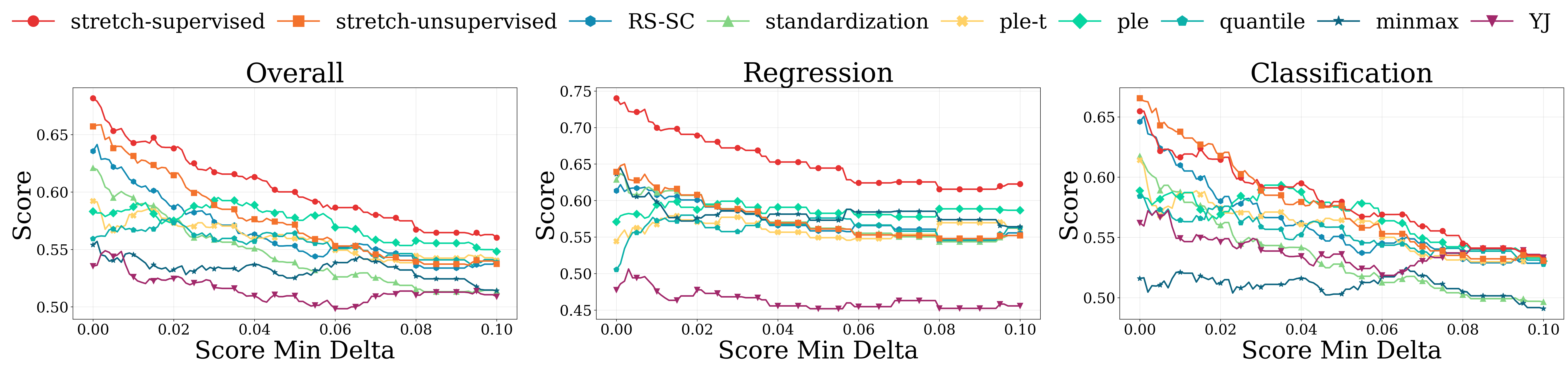}
    \end{center}
    \caption{Sensitivity analysis of significance thresholds. Average normalized score is plotted against a fixed minimum separation $\delta$. Larger $\delta$ values treat small gaps as ties (score 0.5), leading all methods to converge toward 0.5. Unlike the adaptive threshold used in Table~\ref{tab:transformations_performance}, this analysis varies $\delta$ across fixed values. Critically, the relative performance ranking remains stable across all reasonable thresholds well before convergence, confirming robustness to threshold choice. Key takeaways: \n{\textcolor{BrickRed}{supervised stretch} shows the strongest regression performance}, and both \textcolor{BrickRed}{supervised} and \textcolor{orange}{unsupervised stretch} show top-tier performance in classification. \textbf{Abbreviations}: PLE-T (PLE with Tree-based binning), Quantile (Quantile Gaussian), RS-SC (RobustScale+SmoothClip), and YJ (Yeo Johnson).}
    %\caption{Sensitivity analysis showing the robustness of our findings to the choice of significance threshold. The plots show the average normalized score as a function of a minimum performance separation threshold, $\delta$. Unlike the adaptive, statistics-based threshold used for our main results in Table~\ref{tab:transformations_performance}, this analysis explores a range of fixed, absolute values for $\delta$. As $\delta$ increases, more minor performance gaps are treated as ties (score 0.5), causing all methods' scores to converge to 0.5. Critically, the relative performance ranking remains stable across all reasonable thresholds before convergence, confirming that our conclusions are not artifacts of a specific threshold choice. Key takeaways are the clear dominance of \textcolor{BrickRed}{supervised stretch} in regression and the top-tier performance of both \textcolor{BrickRed}{supervised} and \textcolor{orange}{unsupervised stretch} in classification.}
    \label{fig:min-delta}
\end{figure}

\noindent\textbf{Transformation Methods.}
We compare our proposed supervised and unsupervised stretch against seven established baselines: \textbf{standardization} (z-score normalization), \textbf{Yeo-Johnson (YJ)} power transformation \citep{yeo_johnson}, \textbf{quantile transformation} to a Gaussian distribution, \textbf{min-max} scaling to $[0, 1]$, \textbf{RobustScale+SmoothClip (RS-SC)} as originally used in RealMLP \citep{realmlp}, \textbf{Piecewise Linear Encoding (PLE)} \citep{ple}, and \textbf{PLE with Tree-based binning (PLE-T)}. PLE is an unsupervised method that typically uses quantile-based binning, whereas PLE-T is a supervised variant that leverages a tree-based model to create bins informed by the target variable. To our knowledge, PLE-T is the only existing supervised transformation for numeric features in the literature and thus serves as our primary supervised baseline. For transformations that only adjust distribution shape (Yeo-Johnson, quantile, PLE, PLE-T, and stretch variants), we apply standardization afterward to ensure consistent scaling. Implementation details regarding computational constraints are provided in Appendix~\ref{app:fallback-implementation}.

\noindent\textbf{Evaluation Protocol.}
Following the benchmark protocols \citep{talent1,talent2}, we use the official TALENT train, validation, and test partitions. We conduct Bayesian optimization using Optuna \citep{optuna} with 100 trials for each dataset-model-transformation combination, jointly tuning model and transformation hyperparameters (e.g., number of bins for stretch and PLE) using only the training and validation partitions. The test partition is used only after configuration selection. Each selected configuration is evaluated with 15 random seeds on the fixed split; these seeds measure initialization, minibatch-order, and optimization variability, not variability over alternative data splits. We use accuracy for classification and $R^2$ (clipped to $[0, 1]$) for regression as primary metrics.

\noindent\textbf{Aggregation Protocol.}
\n{For aggregate comparisons across heterogeneous datasets we follow \citet{whydotree} and apply per-(dataset, model) min-max normalization to obtain a Normalized Score, paired with an adaptive significance filter that maps statistically inconsequential pairs to a tie ($0.5$); these tie pairs are also excluded from the Avg.\ Acc and Avg.\ $R^2$ columns of Table~\ref{tab:transformations_performance}.} \BV{We also consider head-to-head comparisons of different methods by scrutinizing whether their difference is statistically significant (see Figure~\ref{fig:heatmap}):} \n{$t_1$ wins over $t_2$ only when its performance exceeds $t_2$'s by more than the larger seed standard deviation. The exact formulas, threshold definitions, and filtering rules are given in Appendix~\ref{app:experiment-setup}; the sensitivity analysis in Figure~\ref{fig:min-delta} sweeps a range of fixed thresholds to confirm that the rankings are not artefacts of any particular choice.}

\begin{table}[t]
\centering
\caption{Performance of transformations grouped by model. Normalized Score is computed separately within each (dataset, model) panel and therefore compares transformations within a model; it should not be used to compare model quality across rows. It uses an adaptive significance filter: for each panel we collect the per-method seed standard deviation $\sigma_t$ (computed over $15$ seeds) and use the \emph{median} of $\{\sigma_t\}$ across methods as the threshold; if the gap between the best and worst method is no larger than this median seed std the pair is considered a tie (score $0.5$). Tie pairs are excluded from the `Avg.~Acc' and `Avg.~$R^2$' calculations. The trailing ``$\pm$'' on Avg.~Acc and Avg.~$R^2$ is the standard error of the mean across datasets within the task category, $\sigma_{\rm across\ datasets}/\sqrt{n}$, where $n$ is the number of (dataset, model) pairs that contribute to the cell after tie-exclusion. If any single (dataset, model) pair contains a method whose seed std exceeds $0.20$ (i.e., per-pair seed SEM $> 0.20/\sqrt{15} \approx .05$), that pair is flagged as ``$>\!.05$'' rather than mixed into the per-cell average. For RealMLP, RS-SC denotes its original numeric preprocessing; the other transformation cells disable the internal RS-SC stage while retaining the common hyperparameter-tuning protocol. For performance stratified by dataset feature composition, see Appendix~\ref{sec:performance_breakdown}. \textbf{Abbreviations}: Sup. (Stretch Supervised), Unsup. (Stretch Unsupervised).}
\label{tab:transformations_performance}
\begin{adjustbox}{max width=\textwidth}
\begin{tabular}{c c ccccccccc}
\toprule
\multirow{2.5}{*}{\textbf{Metric}} & \multirow{2.5}{*}{\textbf{Model}} & \multicolumn{9}{c}{\textbf{Transformations}} \\[0.0em]
\cmidrule(lr){3-11}
& & \makebox[1.2cm][c]{\parbox{1.2cm}{\centering\textbf{Sup.}}} & \makebox[1.2cm][c]{\parbox{1.2cm}{\centering\textbf{Unsup.}}} & \makebox[1.2cm][c]{\parbox{1.2cm}{\centering\textbf{Minmax}}} & \makebox[1.2cm][c]{\parbox{1.2cm}{\centering\textbf{PLE}}} & \makebox[1.2cm][c]{\parbox{1.2cm}{\centering\textbf{PLE-T}}} & \makebox[1.2cm][c]{\parbox{1.2cm}{\centering\textbf{Quantile}}} & \makebox[1.2cm][c]{\parbox{1.2cm}{\centering\textbf{RS-SC}}} & \makebox[1.2cm][c]{\parbox{1.2cm}{\centering\textbf{Standard}}} & \makebox[1.2cm][c]{\parbox{1.2cm}{\centering\textbf{YJ}}} \\
\midrule
\multirow{5}{*}{\shortstack{Overall \\ Score}}
    & ftt & \textbf{0.6942} & 0.6543 & 0.3824 & 0.6405 & 0.6461 & 0.6559 & 0.5643 & 0.6214 & 0.5536 \\
    & mlp & 0.6343 & 0.6275 & 0.5600 & \textbf{0.6835} & 0.6182 & 0.5526 & 0.6653 & 0.6423 & 0.5790 \\
    & mlp\_plr & 0.6238 & \textbf{0.7347} & 0.6037 & 0.4513 & 0.5763 & 0.5673 & 0.6091 & 0.6443 & 0.5802 \\
    & realmlp & \textbf{0.7078} & 0.6348 & 0.6321 & 0.5457 & 0.5967 & 0.5566 & 0.6602 & 0.5919 & 0.4482 \\
    & resnet & \textbf{0.7237} & 0.6610 & 0.6175 & 0.5999 & 0.5294 & 0.4854 & 0.6989 & 0.6205 & 0.5160 \\
\midrule
\multirow{5}{*}{\shortstack{Cls. \\ Score}}
    & ftt & 0.6815 & 0.6391 & 0.3434 & 0.6711 & \textbf{0.6929} & 0.6838 & 0.5419 & 0.6003 & 0.5598 \\
    & mlp & 0.5738 & 0.6322 & 0.4900 & \textbf{0.6806} & 0.6324 & 0.5509 & 0.6480 & 0.6525 & 0.5679 \\
    & mlp\_plr & 0.6134 & \textbf{0.7265} & 0.5379 & 0.4578 & 0.5675 & 0.6077 & 0.6473 & 0.6354 & 0.6215 \\
    & realmlp & \textbf{0.7074} & 0.6784 & 0.6799 & 0.5051 & 0.6110 & 0.5871 & 0.6961 & 0.6112 & 0.4656 \\
    & resnet & 0.6809 & 0.6702 & 0.5473 & 0.6180 & 0.5570 & 0.5033 & \textbf{0.7054} & 0.5914 & 0.5763 \\
\midrule
\multirow{5}{*}{\shortstack{Reg. \\ Score}}
    & ftt & \textbf{0.7217} & 0.6872 & 0.4669 & 0.5743 & 0.5446 & 0.5953 & 0.6129 & 0.6671 & 0.5400 \\
    & mlp & \textbf{0.7653} & 0.6173 & 0.7117 & 0.6897 & 0.5874 & 0.5561 & 0.7028 & 0.6203 & 0.6030 \\
    & mlp\_plr & 0.6464 & \textbf{0.7525} & 0.7463 & 0.4372 & 0.5954 & 0.4796 & 0.5262 & 0.6635 & 0.4907 \\
    & realmlp & \textbf{0.7089} & 0.5403 & 0.5287 & 0.6337 & 0.5656 & 0.4907 & 0.5824 & 0.5501 & 0.4104 \\
    & resnet & \textbf{0.8165} & 0.6409 & 0.7697 & 0.5606 & 0.4697 & 0.4464 & 0.6847 & 0.6838 & 0.3855 \\
\midrule
\multirow{5}{*}{Avg. Acc}
    & ftt & \textbf{0.827}{\scriptsize$\pm$.004} & 0.824{\scriptsize$\pm$.003} & 0.805{\scriptsize$\pm$.002} & 0.826{\scriptsize$\pm$.002} & 0.826{\scriptsize$\pm$.002} & \textbf{0.827}{\scriptsize$\pm$.003} & 0.819{\scriptsize$\pm$.003} & 0.816{\scriptsize$\pm$.004} & 0.821{\scriptsize$\pm$.003} \\
    & mlp & 0.821{\scriptsize$\pm$.003} & 0.825{\scriptsize$\pm$.002} & 0.815{\scriptsize$\pm$.002} & \textbf{0.828}{\scriptsize$\pm$.002} & 0.822{\scriptsize$\pm$.002} & 0.816{\scriptsize$\pm$.002} & 0.823{\scriptsize$\pm$.002} & 0.819{\scriptsize$\pm$.002} & 0.822{\scriptsize$\pm$.002} \\
    & mlp\_plr & 0.826{\scriptsize$\pm$.001} & 0.826{\scriptsize$\pm$.002} & 0.821{\scriptsize$\pm$.002} & 0.821{\scriptsize$\pm$.002} & 0.821{\scriptsize$\pm$.003} & \textbf{0.827}{\scriptsize$\pm$.002} & 0.823{\scriptsize$\pm$.002} & 0.821{\scriptsize$\pm$.002} & 0.821{\scriptsize$\pm$.003} \\
    & realmlp & 0.837{\scriptsize$\pm$.002} & 0.838{\scriptsize$\pm$.001} & \textbf{0.839}{\scriptsize$\pm$.002} & 0.833{\scriptsize$\pm$.002} & 0.835{\scriptsize$\pm$.002} & 0.837{\scriptsize$\pm$.002} & 0.838{\scriptsize$\pm$.002} & 0.833{\scriptsize$\pm$.001} & 0.834{\scriptsize$\pm$.002} \\
    & resnet & \textbf{0.826}{\scriptsize$\pm$.002} & 0.824{\scriptsize$\pm$.003} & 0.815{\scriptsize$\pm$.003} & 0.824{\scriptsize$\pm$.002} & 0.823{\scriptsize$\pm$.002} & 0.816{\scriptsize$\pm$.002} & 0.825{\scriptsize$\pm$.002} & 0.819{\scriptsize$\pm$.002} & 0.820{\scriptsize$\pm$.002} \\
\midrule
\multirow{5}{*}{Avg. $R^2$}
    & ftt & \textbf{0.691}{\scriptsize$\pm$.002} & 0.681{\scriptsize$\pm$.002} & 0.666{\scriptsize$\pm$.003} & 0.680{\scriptsize$\pm$.006} & 0.667{\scriptsize$\pm$.003} & 0.681{\scriptsize$\pm$.002} & 0.673{\scriptsize$\pm$.003} & 0.682{\scriptsize$\pm$.001} & 0.667{\scriptsize$\pm$.001} \\
    & mlp & \textbf{0.630}{\scriptsize$\pm$.005} & 0.606{\scriptsize$\pm$.007} & 0.628{\scriptsize$\pm$.008} & 0.608{\scriptsize$\pm$.039} & 0.588{\scriptsize$\pm$.011} & 0.619{\scriptsize$\pm$.005} & 0.619{\scriptsize$\pm$.005} & 0.591{\scriptsize$\pm$.042} & 0.545{\scriptsize$\pm$\,$>$\,.05} \\
    & mlp\_plr & 0.698{\scriptsize$\pm$.004} & 0.696{\scriptsize$\pm$.004} & 0.659{\scriptsize$\pm$.004} & 0.686{\scriptsize$\pm$.002} & \textbf{0.702}{\scriptsize$\pm$.002} & 0.669{\scriptsize$\pm$.005} & 0.681{\scriptsize$\pm$.004} & 0.675{\scriptsize$\pm$.003} & 0.649{\scriptsize$\pm$.005} \\
    & realmlp & \textbf{0.736}{\scriptsize$\pm$.004} & 0.685{\scriptsize$\pm$.004} & 0.679{\scriptsize$\pm$.003} & \textbf{0.736}{\scriptsize$\pm$.003} & 0.723{\scriptsize$\pm$.003} & 0.707{\scriptsize$\pm$.005} & 0.689{\scriptsize$\pm$.004} & 0.686{\scriptsize$\pm$.002} & 0.683{\scriptsize$\pm$.008} \\
    & resnet & 0.651{\scriptsize$\pm$.003} & 0.631{\scriptsize$\pm$.006} & \textbf{0.655}{\scriptsize$\pm$.002} & 0.639{\scriptsize$\pm$\,$>$\,.05} & 0.631{\scriptsize$\pm$.005} & 0.625{\scriptsize$\pm$.011} & 0.640{\scriptsize$\pm$.002} & 0.641{\scriptsize$\pm$.003} & 0.542{\scriptsize$\pm$\,$>$\,.05} \\
\bottomrule
\end{tabular}
\end{adjustbox}
\end{table}

\subsection{Results and Analysis}\label{sec:analysis}

Our analysis draws on three complementary perspectives. First, Figure~\ref{fig:min-delta} presents a sensitivity analysis using a range of fixed significance thresholds ($\delta$) to demonstrate the robustness of our findings. Second, Table~\ref{tab:transformations_performance} provides a detailed breakdown for each model using an adaptive threshold based on the median standard error of each specific evaluation. This reveals how different architectures interact with each transformation under a statistically grounded criterion. Finally, the pairwise win rate heatmap in Figure~\ref{fig:heatmap} offers direct head-to-head comparisons to identify the most consistently superior methods. Together, these analyses lead to \n{four} key findings:

\begin{figure}[t]
    \begin{center}
    \includegraphics[width=\linewidth]{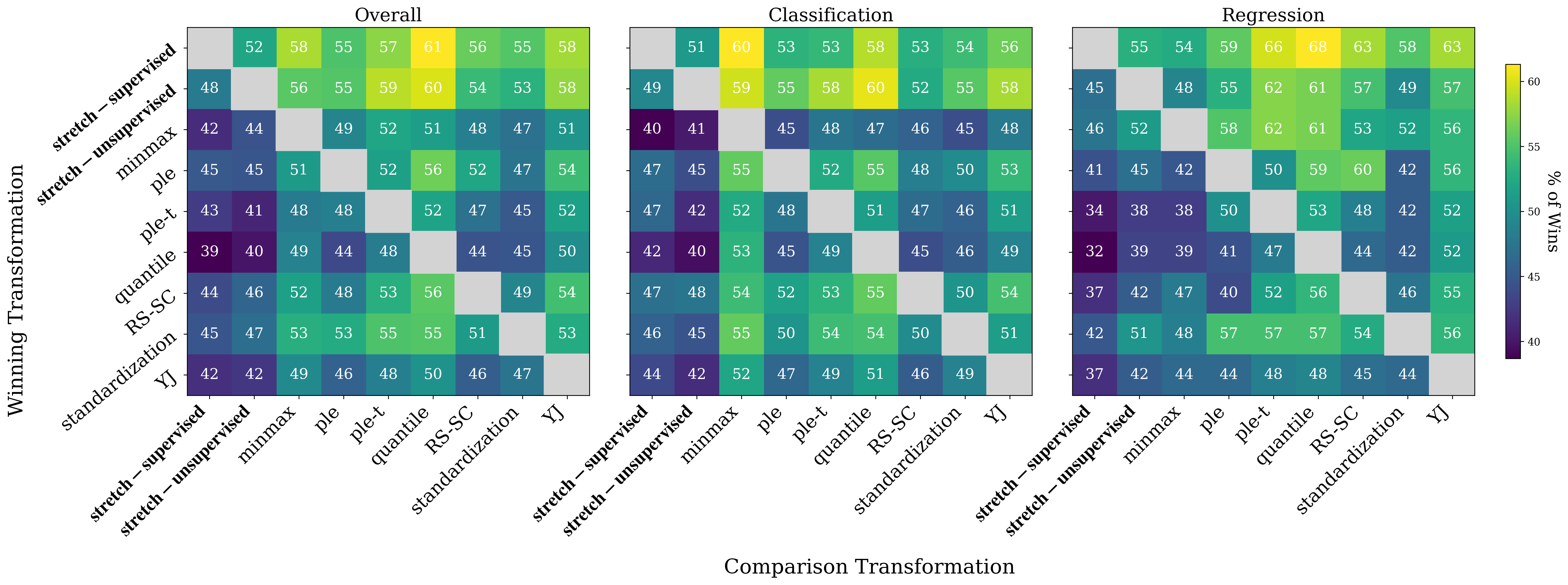}
    \end{center}
    \caption{Pairwise win-rates between transformations, grouped by task. Each cell $(i, j)$ shows the percentage of times transformation $i$ (row) statistically outperforms transformation $j$ (column). \n{Supervised stretch has the strongest aggregate pairwise performance across methods and tasks. Unsupervised stretch is the strongest unsupervised baseline under this criterion}, second only to its supervised counterpart, with particularly competitive performance in classification. For detailed win-loss-tie analysis against the strongest unsupervised (PLE, Standardization) and supervised (PLE-T) baselines, see Appendix~\ref{app:pairwise_scatter}.}
    \label{fig:heatmap}
\end{figure}

\n{\textbf{1. Supervised stretch achieves the strongest aggregate performance, especially in regression.}} \n{Across all evaluations, supervised stretch stands out as the strongest method, with its advantage most pronounced in regression, where it maintains a substantial margin over all competitors across significance thresholds $\delta$ (Figure~\ref{fig:min-delta}, middle); the competition is tighter in classification but it remains in the top tier (model-specific scores in Table~\ref{tab:transformations_performance}).} The win-rate analysis (Figure~\ref{fig:heatmap}) further shows that it statistically outperforms every alternative \n{under the pairwise criterion}, \n{with decisive regression win-loss records of \textbf{18-7} against PLE, \textbf{28-9} against PLE-T, and \textbf{14-4} against standardization (Appendix~\ref{app:pairwise_scatter}).} \n{These results support our hypothesis that smoothing target functions via Dirichlet energy minimization improves neural network performance, and the method is not fragile under label noise: with label flipping (classification) and additive Gaussian target noise (regression), supervised stretch keeps a competitive mean rank across noise levels and is top-ranked at the highest level tested (Appendix~\ref{app:noise_analysis}).}

\n{\textbf{2. Unsupervised stretch is a strong target-agnostic method.}} \n{Unsupervised stretch consistently ranks as a strong runner-up: it outperforms all methods except its supervised counterpart under the aggregate pairwise comparison (\Cref{fig:heatmap}) and is especially competitive in classification (\Cref{fig:min-delta}, right), with classification win-loss records of \textbf{38--25} vs.\ PLE, \textbf{38--18} vs.\ PLE-T, and \textbf{26--13} vs.\ Standardization (\Cref{app:pairwise_scatter}).} Its per-feature memory overhead is $O(1)$ compared to PLE's $O(T)$, avoiding computational bottlenecks that can make PLE impractical for large datasets. 

\textbf{3. Transformation preferences vary by architecture and task type.} \n{Although supervised stretch is the strongest generalist, architectural preferences remain (e.g., unsupervised stretch is the top choice for MLP-PLR). More importantly, task type drives the largest differences: regression benefits more from advanced transformations than classification, with supervised stretch yielding the largest gains in continuous prediction tasks (\Cref{fig:min-delta}, middle). This suggests that explicitly engineering smoother feature-to-target mappings is especially valuable for regression, an underexplored direction in a literature largely focused on classification.}

\textbf{4. Non-uniform marginal signals help explain why smoothing is beneficial, especially for regression.} Our marginal analysis (Appendix~\ref{app:marginal_snalysis}) shows that real-world tabular features exhibit non-uniform marginal distributions with localized regions of sharp variation (high relative marginal slopes); this structural heterogeneity creates high-frequency components that are difficult for neural networks to learn due to spectral bias. Supervised stretch explicitly identifies and smooths these high-gradient regions, which is particularly relevant for regression with continuous targets and helps explain why our gains are larger there than in classification, where only a decision boundary is needed. \looseness=-1

\section{Conclusion}\label{sec:conclusion}

\n{This paper introduced the \textbf{stretch transformation framework}, a principled approach to numeric feature preprocessing for tabular deep learning that frames transformation as Dirichlet-energy minimization of the target function. The framework yields two methods: \textbf{supervised stretch}, which systematically exploits target information to create smoother target functions and achieves the strongest empirical performance, and \textbf{unsupervised stretch}, which maximizes worst-case separation through uniform density redistribution.} \n{Our analysis also explains existing techniques: unsupervised stretch clarifies why empirical CDF transformation improves learning despite being target-agnostic and reveals its equivalence to PLE via arc-length parameterization, while supervised stretch connects to target encoding through shared use of target variation, with the number of bins providing principled regularization.} \n{Empirically, supervised stretch obtains consistent gains across 38 datasets, supporting the hypothesis that target-aware transformation enhances neural network performance and challenging the prevailing paradigm of unsupervised preprocessing in tabular deep learning.}

\textbf{Limitations and Future Work.} Our framework's design could be enhanced by exploring smoother, spline-based alternatives, and adaptive binning strategies. The current transformation is marginal and therefore does not optimize a joint or conditional smoothness objective; the downstream model may learn feature interactions, but the preprocessing itself is not interaction-aware. Theoretically, the analysis could be generalized beyond the small-$\sigma$ RBF kernel approximation. Applying Stretch as an option in a TabPFN inference-time preprocessing ensemble is also an interesting direction that is outside the controlled architecture study here.

\bibliography{neurips2026}
\bibliographystyle{plainnat}

\newpage

\appendix

\section{Theoretical Foundations and Derivations}
\label{app:derivations}

\subsection{General Objective: From Kernel Smoothness to Dirichlet Energy}\label{app:dirichlet-motivation}

\paragraph{Motivation via Heat Kernel Expansion.}
The primary goal is to find a transformation that makes the target function smoother in the transformed space. We quantify smoothness using kernel correlation with a Gaussian RBF kernel, $k_\sigma(y,z) = \exp(-(y-z)^2/(2\sigma^2))$. The integral operator $T_\sigma$ associated with this kernel can be expressed via the heat kernel expansion as $T_\sigma = e^{\frac{\sigma^2}{2}\partial_{yy}}$, where $\partial_{yy}$ is the second derivative operator \citep{varadhan1967behavior,molchanov1975diffusion}. For a small bandwidth $\sigma$, a Taylor expansion gives $T_\sigma = I + \frac{\sigma^2}{2}\partial_{yy} + O(\sigma^4)$.

To isolate the smoothness-dependent term, we consider the centered kernel correlation:
\begin{align}
    \langle g, (T_\sigma - I)g\rangle &= \left\langle g, \left(\frac{\sigma^2}{2}\partial_{yy} + O(\sigma^4)\right)g\right\rangle = \frac{\sigma^2}{2}\int_0^1 g(y) \cdot g''(y)\,dy + O(\sigma^4).
\end{align}
Using integration by parts (assuming Neumann boundary conditions, $g'(0) = g'(1) = 0$), this simplifies to:
\begin{align}
    \int_0^1 g(y) \cdot g''(y)\,dy &= \left[g(y) \cdot g'(y)\right]_0^1 - \int_0^1 g'(y) \cdot g'(y)\,dy = - \int_0^1 \|g'(y)\|_2^2\,dy.
\end{align}
Therefore, maximizing kernel smoothness in this regime is equivalent to minimizing the Dirichlet energy:
\begin{equation}
    \langle g, (T_\sigma - I)g\rangle = -\frac{\sigma^2}{2}\int_0^1 \|g'(y)\|_2^2\,dy + O(\sigma^4), \quad \text{so we minimize } \mathcal{E}[g] = \int_0^1 \|g'(y)\|_2^2\,dy.
\end{equation}
\n{For vector-valued targets, this identity is applied componentwise and summed over the target dimensions.}

\paragraph{Discretization of Dirichlet Energy.}
For a dataset $\{x_i\}_{i=1}^n$ (assumed sorted) with transformed values $y_i = s(x_i)$, we approximate the continuous target function $g(y)$ by its piecewise linear interpolant, denoted by $\tilde g$. The interpolant $\tilde g$ is constructed by connecting the points $(y_i, f(x_i))$. By the definition of our transformation, the true function value at the transformed point $y_i$ is precisely $g(y_i) = f(s^{-1}(y_i)) = f(x_i)$.

The derivative of this interpolant, $\tilde g'(y)$, is constant over any interval $[y_i, y_{i+1}]$ and is given by:
$$ \tilde g'(y) = \frac{f(x_{i+1}) - f(x_i)}{y_{i+1} - y_i} = \frac{\Delta f_i}{\Delta y_i}. $$
The total discrete Dirichlet energy is the integral of the squared norm of this derivative, which simplifies to a sum over the intervals:
\begin{equation}
    \mathcal{E}_{\mathrm{disc}} = \sum_{i=1}^{n-1} \int_{y_i}^{y_{i+1}} \|\tilde{g}'(y)\|_2^2\,dy = \sum_{i=1}^{n-1} \frac{\|\Delta f_i\|_2^2}{\Delta y_i}.
\end{equation}
This discrete energy serves as the foundation for both our unsupervised and supervised methods.

\subsection{Unsupervised Stretch: Derivation of the Minimax Objective}\label{app:unsupervised-objective-derivation}

\paragraph{Worst-Case Dirichlet Energy Formulation.}
Without target information, we formulate a minimax problem to find the spacing allocation that minimizes the worst-case Dirichlet energy. The optimization is performed over all possible target functions $f$ with \n{bounded local variation, i.e., $\|\Delta f_i\|_2^2 \leq C$ for all $i$}. Since the discrete Dirichlet energy $\mathcal{E}_{\mathrm{disc}} = \sum_{i=1}^{n-1}\frac{\|\Delta f_i\|_2^2}{\Delta y_i}$ depends only on the sequence of difference vectors $\{\Delta f_i\}_{i=1}^{n-1}$, this is equivalent to solving the following problem:
\begin{equation}
    \max_{\{\Delta f_i\}} \sum_{i=1}^{n-1}\frac{\|\Delta f_i\|_2^2}{\Delta y_i}
    \quad \text{s.t.} \quad \|\Delta f_i\|_2^2 \leq C, \ \forall i.
\end{equation}
\n{This bounded local variation constraint assumes that for an unknown function, the magnitude of its change over any adjacent interval is bounded, which is a practical assumption for unsupervised scenarios where we want to be robust against arbitrary local fluctuations.}

\n{\paragraph{Solving the Minimax Problem.}
The inner maximization is straightforward. Since each term $\frac{\|\Delta f_i\|_2^2}{\Delta y_i}$ is nonnegative and increasing in $\|\Delta f_i\|_2^2$, the worst case is attained at the upper bound:
\begin{equation}
    \|\Delta f_i\|_2^2 = C, \quad \forall i.
\end{equation}
Substituting this into the objective yields:
\begin{equation}
    \max_{\{\Delta f_i\}} \sum_{i=1}^{n-1}\frac{\|\Delta f_i\|_2^2}{\Delta y_i}
    = C\sum_{i=1}^{n-1}\frac{1}{\Delta y_i}.
\end{equation}
Since $C$ is a positive constant, the outer minimization over spacings becomes:
\begin{equation}
    \min_{\{\Delta y_i > 0\}} \sum_{i=1}^{n-1}\frac{1}{\Delta y_i}
    \quad \text{s.t.} \quad \sum_{i=1}^{n-1}\Delta y_i = 1.
\end{equation}

\paragraph{Uniform Spacing Solution.}
This is a convex optimization problem. The Lagrangian is:
\begin{equation}
    \mathcal{L}(\{\Delta y_i\},\lambda)
    = \sum_{i=1}^{n-1}\frac{1}{\Delta y_i}
    + \lambda\left(\sum_{i=1}^{n-1}\Delta y_i - 1\right).
\end{equation}
Taking derivative with respect to $\Delta y_i$ and setting it to zero gives:
\begin{equation}
    -\frac{1}{(\Delta y_i)^2} + \lambda = 0,
\end{equation}
which implies that all $\Delta y_i$ are equal. Using the constraint $\sum_{i=1}^{n-1}\Delta y_i = 1$, we obtain:
\begin{equation}
    \Delta y_i^* = \frac{1}{n-1}, \quad \forall i.
\end{equation}
Thus, the minimax-optimal spacing allocation is uniform.}

\subsection{Derivation of Supervised Objective}\label{app:supervised-objective-derivation}

\paragraph{Exact Piecewise-Linear Objective.}
When target information is available, the discrete Dirichlet energy from Appendix~\ref{app:dirichlet-motivation} can be written explicitly under the piecewise-linear stretch map. Consider bin $B_t=[b_{t-1},b_t]$ with allocated width $w_t$. For adjacent samples $x_i,x_{i+1}\in B_t$, the transformation is linear within the bin, so
\begin{equation}
    \Delta y_i
    = y_{i+1}-y_i
    = \frac{w_t}{b_t-b_{t-1}}\Delta x_i,
\end{equation}
where $\Delta x_i=x_{i+1}-x_i$. Substituting this into the discrete Dirichlet energy gives the exact within-bin contribution
\begin{equation}
    E_t
    =
    \frac{b_t-b_{t-1}}{w_t}
    \sum_{i\in I_t}
    \frac{\|\Delta f_i\|_2^2}{\Delta x_i},
\end{equation}
where $I_t=\{i:x_i,x_{i+1}\in B_t\}$. This exact objective is sensitive to very small input spacings: finite-sample noise in estimated target differences can make this term unstable when $\Delta x_i$ is very small.

\paragraph{Robust Lower-Bound Surrogate.}
To obtain a more stable bin-level objective, we use a lower bound based on the total target variation within each bin. Define
\begin{equation}
    S_t = \sum_{i\in I_t}\|\Delta f_i\|_2.
\end{equation}
Since the allocated width of bin $t$ satisfies $\sum_{i\in I_t}\Delta y_i=w_t$, Cauchy--Schwarz gives
\begin{equation}
    \left(\sum_{i\in I_t}\|\Delta f_i\|_2\right)^2
    \leq
    \left(\sum_{i\in I_t}\frac{\|\Delta f_i\|_2^2}{\Delta y_i}\right)
    \left(\sum_{i\in I_t}\Delta y_i\right).
\end{equation}
Therefore,
\begin{equation}
    \sum_{i\in I_t}\frac{\|\Delta f_i\|_2^2}{\Delta y_i}
    \geq
    \frac{S_t^2}{w_t}.
\end{equation}
We adopt this robust lower bound as our objective:
\begin{equation}
    \min_{\{w_t>0\}} \sum_{t=1}^T \frac{S_t^2}{w_t}
    \quad \text{s.t.} \quad \sum_{t=1}^T w_t=1.
\end{equation}
This objective is a robust bin-level surrogate rather than the exact unrestricted sample-spacing optimum; equality would additionally require the within-bin transformed spacings to align with target variation. The bin count $T$ controls the corresponding trade-off: smaller $T$ preserves more local structure, whereas larger $T$ allows finer, potentially noisier, width reallocation. Solving the Lagrangian yields
\begin{equation}
    w_t^\star = \frac{S_t}{\sum_{u=1}^T S_u}.
\end{equation}

\section{Implementation and Experimental Details}
\label{app:implementation-and-experiments}

\subsection{Out-of-Fold Kernel Regression for Supervised Stretch}
\label{app:oof-implementation}
A naive implementation of supervised stretch would use the entire training set to estimate the target function $f(x)$ needed to compute the bin widths $\{w_t\}$. This, however, would introduce significant information leakage. The transformation $s$ would be designed using the same target values $t_i$ that the downstream model $g_\theta$ is later trained to predict. In essence, the target information would be used twice, once to shape the feature space, and again to train the model within that space. This can lead to overly optimistic performance estimates and poor generalization.

To prevent this, we employ a standard K-fold cross-validation scheme to obtain out-of-fold (OOF) estimates of the marginal regression function, $\widehat{f}(x_i)$, for each sample. \n{This ensures that $t_i$ is not used to construct its own out-of-fold estimate $\hat f(x_i)$. The final bin widths are then computed from OOF predictions across the training set, reducing direct self-label leakage while still using target information at the transformation-design stage.}

For classification, we use stratified K-fold ($K=10$, or fewer if a class is too small) on the one-hot encoded targets. For regression, we use standard K-fold. Within each fold, we fit an adaptive Nadaraya-Watson kernel regressor on the training portion of the data to generate predictions for the validation portion~\citep{KernelRegression, KernelRegression2}. Let $\mathcal{D}_{\mathrm{tr}}$ denote the set of indices of the samples in the training portion for a given fold. The conditional expectation at a point $x$ in the validation set is then estimated as:
\begin{equation}
    \widehat{f}(x) = \frac{\sum_{j \in \mathcal{D}_{\mathrm{tr}}} K_h(x, x_j) \cdot Y_j}{\sum_{j \in \mathcal{D}_{\mathrm{tr}}} K_h(x, x_j) + \epsilon}, \quad \text{where} \quad K_h(x, x_j) = \exp\left(-\frac{(x-x_j)^2}{2h(x)^2}\right).
\end{equation}
Here, $Y_j$ is the target value for the $j$-th sample, and the adaptive bandwidth $h(x)$ is determined by the distance to the $k$-th nearest neighbor of $x$ within the training data, allowing the smoothness of the estimate to vary with local sample density.

After obtaining OOF predictions for all samples, we sort them by the feature value and compute the total variation $S_t$ of these predictions within each quantile bin. Several fallback mechanisms ensure robustness: if a feature has too few unique values, or if the total target variation is negligible, we revert to simpler transformations (identity or unsupervised stretch).

\subsection{Computational Fallback Scheme for PLE}
\label{app:fallback-implementation}
PLE expands feature dimensionality from $d$ to $d \times T$, which can become computationally prohibitive. In our experiments, if the resulting tensor size exceeds a memory limit, we use a fallback scheme that substitutes standardization for PLE on that specific dataset-model combination. This affects fewer than 5\% of the 190 combinations and does not impact our proposed methods.

\subsection{Experimental Setup Details}
\label{app:experiment-setup}
\paragraph{Dataset Selection.}
From the original TALENT Benchmark 1 collection of 42 datasets, we exclude four datasets that contain only categorical features (Amazon\_employee\_access, BNG(tic-tac-toe), led24, splice), yielding our final experimental suite of 38 datasets.

\paragraph{Categorical Feature Handling.} We use indices encoding as the default. FT-Transformer learns embeddings, RealMLP internally converts to one-hot, and other models use the indices directly. For feature-expanding transformations like PLE, FT-Transformer applies a linear projection to each expanded dimension to create tokens.

\paragraph{Hyperparameter Search Spaces.} We conduct Bayesian optimization using Optuna with 100 trials per configuration. The search spaces for models and transformations are detailed in Table~\ref{tab:model_hparams} and Table~\ref{tab:transform_hparams}. We follow the hyperparameter search space design from the original TALENT benchmark~\citep{talent1, talent2}, including their special syntax for sampling methods, which we define here for clarity:
\begin{itemize}[leftmargin=1.5em, itemsep=0pt, topsep=2pt]
    \item \texttt{?distribution[default, ...args]}: An ``optional'' parameter. With $50\%$ probability, the \texttt{default} value is used (e.g., 0.0 for dropout, disabling it). Otherwise, a value is sampled from the specified \texttt{distribution} using the remaining arguments.
    \item \texttt{\$name[...]}: A special, complex sampling function defined by the benchmark's codebase. For instance, \texttt{\$mlp\_d\_layers} samples the number of layers and then the width of each layer within given bounds, with different sampling strategies for the first, middle, and last layers.
\end{itemize}

\begin{table}[h]
\centering
\caption{Transformation hyperparameter search spaces.}
\label{tab:transform_hparams}
\small
\begin{tabular}{ll}
\toprule
\textbf{Transformation} & \textbf{Parameter Search Space} \\
\midrule
Supervised/Unsupervised Stretch & n\_bins: \texttt{?categorical[1, 2, 4, 8, 16, 32,} \\
& \texttt{64, 128, 256, 512, 100000]}$^\dagger$ \\
PLE & n\_bins: \texttt{int[2, 256]} \\
Other transformations & No hyperparameters \\
\bottomrule
\end{tabular}
\end{table}

\begin{table}[h]
\centering
\caption{Model hyperparameter search spaces.}
\label{tab:model_hparams}
\small
\begin{tabular}{lll}
\toprule
\textbf{Model} & \textbf{Parameter} & \textbf{Search Space} \\
\midrule
\multirow{4}{*}{MLP}
& d\_layers & \texttt{\$mlp\_d\_layers[1, 8, 64, 2048]}$^*$ \\
& dropout & \texttt{?uniform[0.0, 0.0, 0.5]}$^\dagger$ \\
& lr & \texttt{loguniform[1e-5, 0.01]} \\
& weight\_decay & \texttt{?loguniform[0.0, 1e-6, 0.001]}$^\dagger$ \\
\midrule
\multirow{7}{*}{MLP-PLR}
& d\_layers & \texttt{\$mlp\_d\_layers[1, 8, 64, 1024]}$^*$ \\
& dropout & \texttt{?uniform[0.0, 0.0, 0.5]}$^\dagger$ \\
& n\_frequencies & \texttt{int[16, 96]} \\
& frequency\_scale & \texttt{loguniform[0.01, 100.0]} \\
& d\_embedding & \texttt{int[16, 64]} \\
& lr & \texttt{loguniform[1e-5, 0.01]} \\
& weight\_decay & \texttt{?loguniform[0.0, 1e-6, 0.001]}$^\dagger$ \\
\midrule
\multirow{8}{*}{FT-Transformer}
& n\_layers & \texttt{int[1, 4]} \\
& d\_token & \texttt{categorical[8, 16, 32, 64, 128]} \\
& residual\_dropout & \texttt{?uniform[0.0, 0.0, 0.2]}$^\dagger$ \\
& attention\_dropout & \texttt{uniform[0.0, 0.5]} \\
& ffn\_dropout & \texttt{uniform[0.0, 0.5]} \\
& d\_ffn\_factor & \texttt{uniform[0.67, 2.67]} \\
& lr & \texttt{loguniform[1e-5, 0.001]} \\
& weight\_decay & \texttt{loguniform[1e-6, 0.001]} \\
\midrule
\multirow{8}{*}{RealMLP}
& num\_emb\_type & \texttt{categorical[none, pbld, pl, plr]} \\
& add\_front\_scale & \texttt{categorical[true, false]} \\
& hidden\_sizes & \texttt{categorical[[256,256,256],} \\
&  & \texttt{[64,64,64,64,64], [512]]} \\
& p\_drop & \texttt{categorical[0.0, 0.15, 0.30]} \\
& act & \texttt{categorical[selu, relu, mish]} \\
& lr & \texttt{loguniform[0.02, 0.3]} \\
& wd & \texttt{categorical[0.0, 0.02]} \\
& ls\_eps & \texttt{categorical[0.0, 0.1]} \\
& plr\_sigma & \texttt{loguniform[0.05, 0.5]} \\
\midrule
\multirow{5}{*}{ResNet}
& n\_layers & \texttt{int[1, 8]} \\
& d & \texttt{int[64, 512]} \\
& d\_hidden\_factor & \texttt{uniform[1.0, 4.0]} \\
& hidden\_dropout & \texttt{uniform[0.0, 0.5]} \\
& residual\_dropout & \texttt{?uniform[0.0, 0.0, 0.5]}$^\dagger$ \\
& lr & \texttt{loguniform[1e-5, 0.01]} \\
& weight\_decay & \texttt{?loguniform[0.0, 1e-6, 0.001]}$^\dagger$ \\
\bottomrule
\end{tabular}
\end{table}

\subsection{Aggregation and Pairwise Comparison Protocol}
\label{app:aggregation-protocol}

This subsection makes the protocol summarised in Section~\ref{sec:experiments} fully explicit. Let $d$, $m$, and $t$ index datasets, models, and transformations.

\paragraph{Per-(dataset, model) min-max normalisation.} Following \citet{whydotree}, we map raw metrics to a normalised score
\begin{equation}\label{eq:appendix-normalized-score}
    \text{score}_{d,m,t} = \frac{\text{metric}_{d,m,t} - \min_t \text{metric}_{d,m,t}}{\max_t \text{metric}_{d,m,t} - \min_t \text{metric}_{d,m,t}},
\end{equation}
so that the best transformation on each $(d,m)$ pair scores $1$ and the worst scores $0$.

\paragraph{Adaptive significance filter.} For each $(d,m)$ we compute the per-method seed standard deviation $\sigma_{d,m,t}$ over the $15$ evaluation seeds and set the threshold
$\delta_{d,m} = \mathrm{median}_t\, \sigma_{d,m,t}$. If the gap $\max_t \text{metric}_{d,m,t} - \min_t \text{metric}_{d,m,t}$ does not exceed $\delta_{d,m}$, the choice of transformation is deemed inconsequential, every $\text{score}_{d,m,t}$ is set to $0.5$, and the pair is excluded from the Avg.\ Acc / Avg.\ $R^2$ averages of Table~\ref{tab:transformations_performance}. The Normalized Score columns of Table~\ref{tab:transformations_performance} use this adaptive threshold; the sensitivity analysis in Figure~\ref{fig:min-delta} replaces $\delta_{d,m}$ with a range of fixed values to confirm that the ranking is robust.

\paragraph{Pairwise win criterion.} For the head-to-head heatmap (Figure~\ref{fig:heatmap}), a transformation $t_1$ is declared to win against $t_2$ on a given $(d,m)$ only if
\begin{equation}\label{eq:appendix-win-criterion}
    \text{metric}_{d,m,t_1} > \text{metric}_{d,m,t_2} + \max\bigl(\sigma_{d,m,t_1}, \sigma_{d,m,t_2}\bigr).
\end{equation}
This requires the gap to exceed the larger of the two seed standard deviations, so that an isolated lucky seed cannot trigger a win. Pairs that do not satisfy~\eqref{eq:appendix-win-criterion} in either direction are recorded as ties and contribute $0.5$ to each side's count. The win-rate reported in the heatmap is the win count divided by the $190$ dataset-model combinations (pairs where both methods are missing are excluded).

%\clearpage

\section{Detailed Pairwise Performance Analysis}
\label{app:pairwise_scatter}

To provide a more granular analysis of the stretch transformation's performance, this section extends the summary presented in the win-rate heatmap \Cref{sec:experiments} (Figure~\ref{fig:heatmap}). The pairwise comparison uses the strict win-loss-tie criterion stated in Eq.~\eqref{eq:appendix-win-criterion}; the win-rate percentages shown in the heatmap are summary statistics derived directly from the raw W-L-T counts across all 190 dataset-model combinations.

\begin{figure}[htbp]
    \centering
    \includegraphics[width=0.46\linewidth]{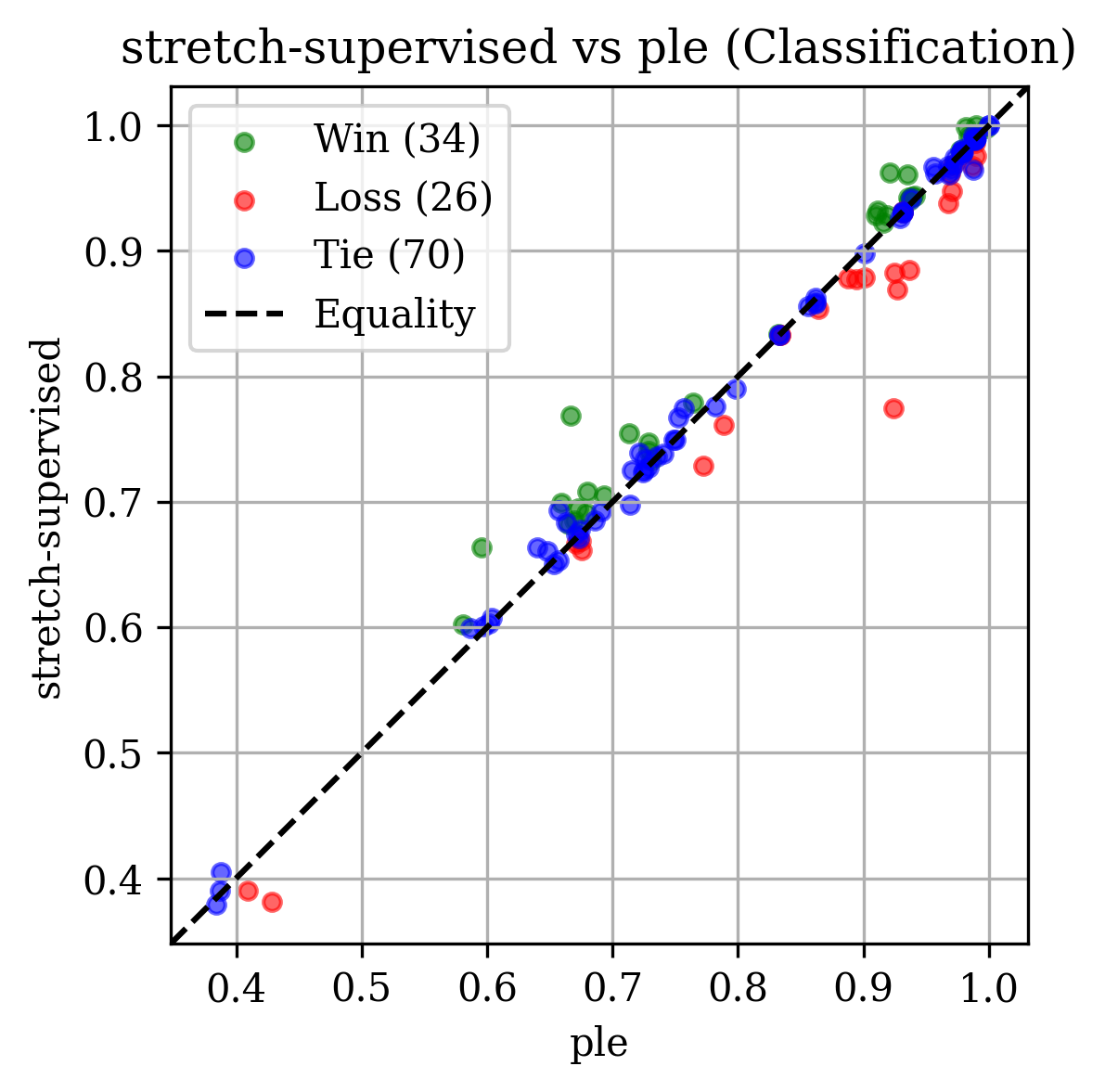}
    \hfill
    \includegraphics[width=0.46\linewidth]{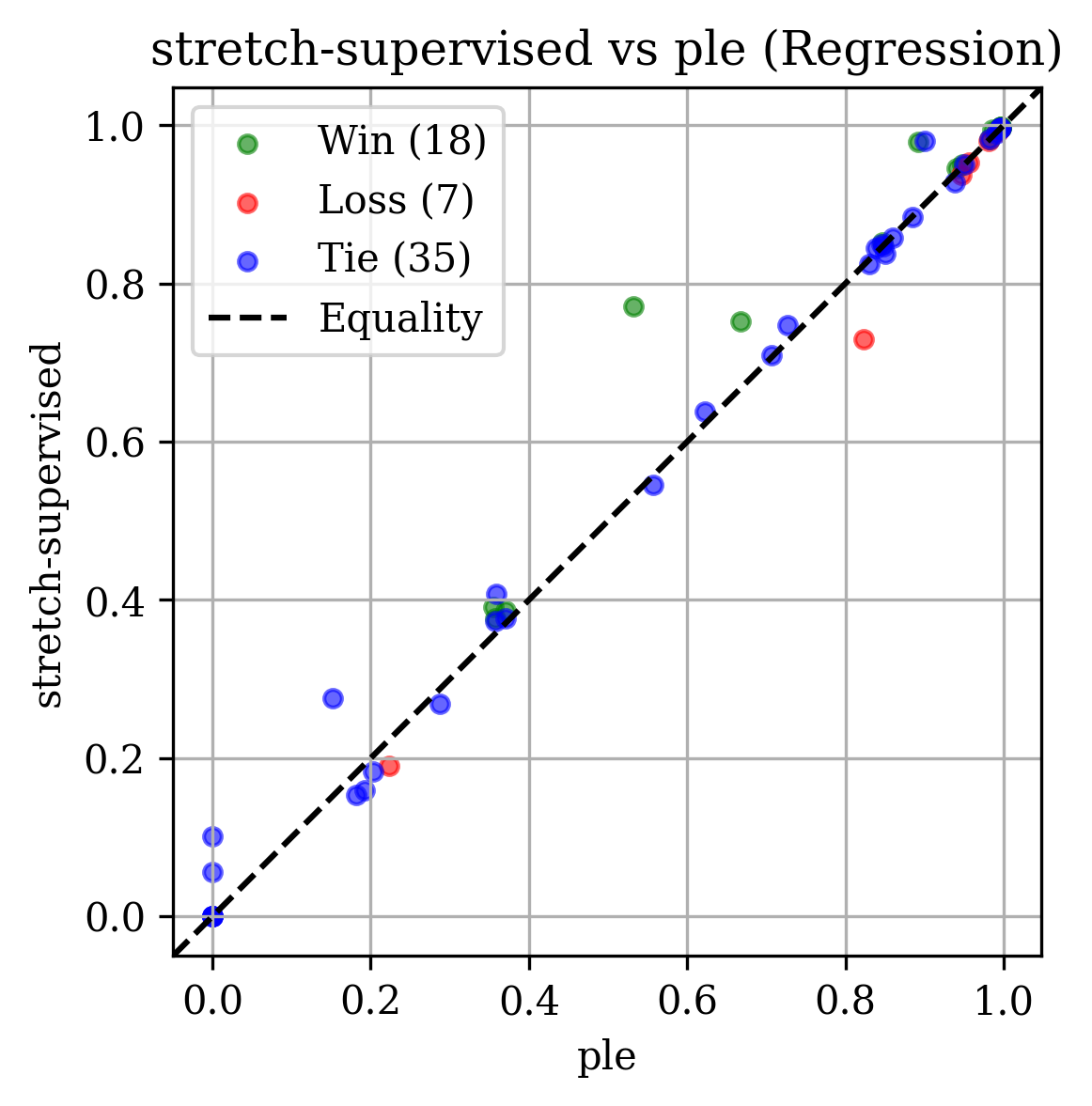}
    \caption{Stretch Supervised vs PLE}
    \label{fig:pairwise_scatter1}
\end{figure}

\begin{figure}[htbp]
    \centering
    \includegraphics[width=0.465\linewidth]{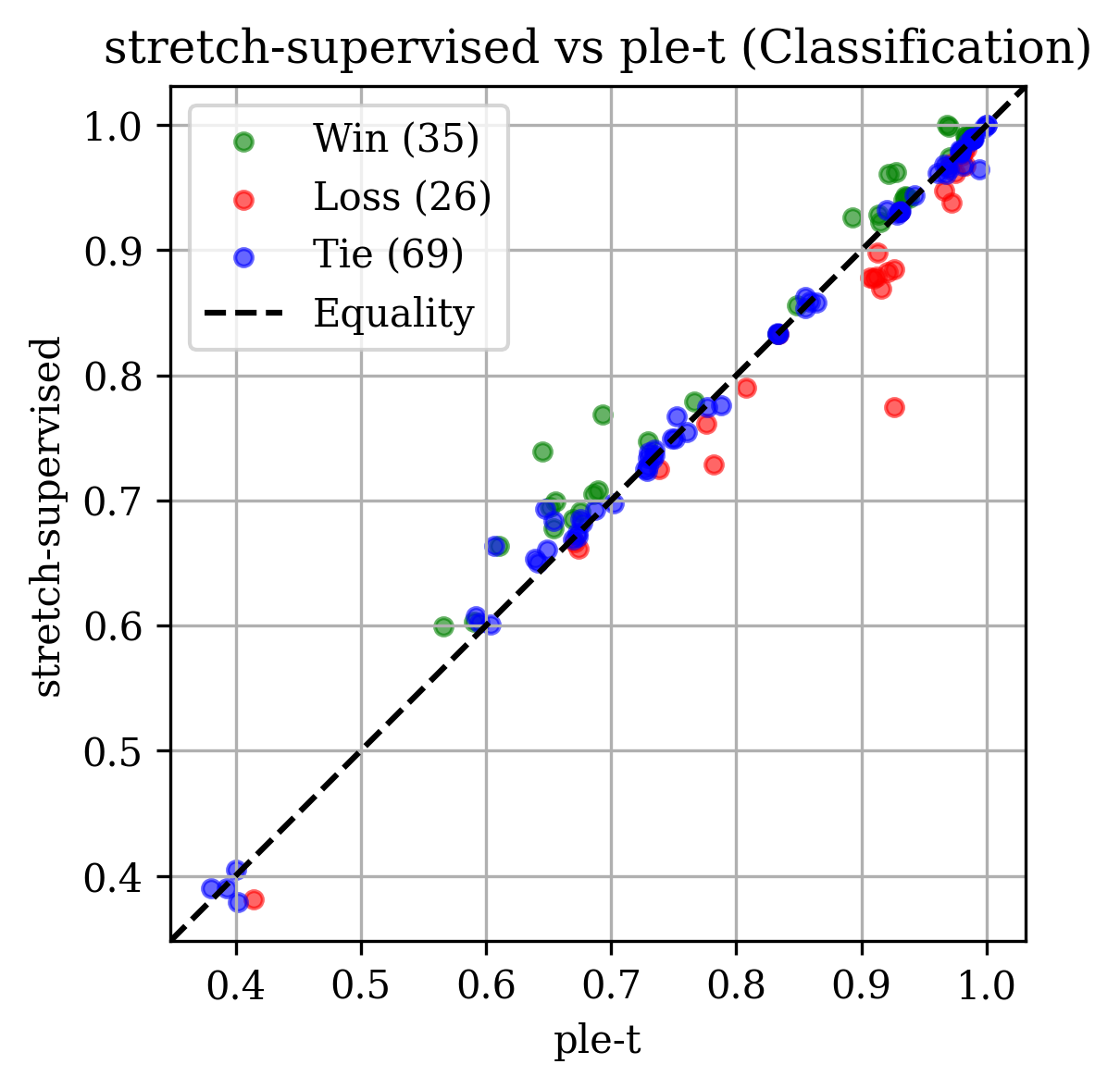}
    \hfill
    \includegraphics[width=0.46\linewidth]{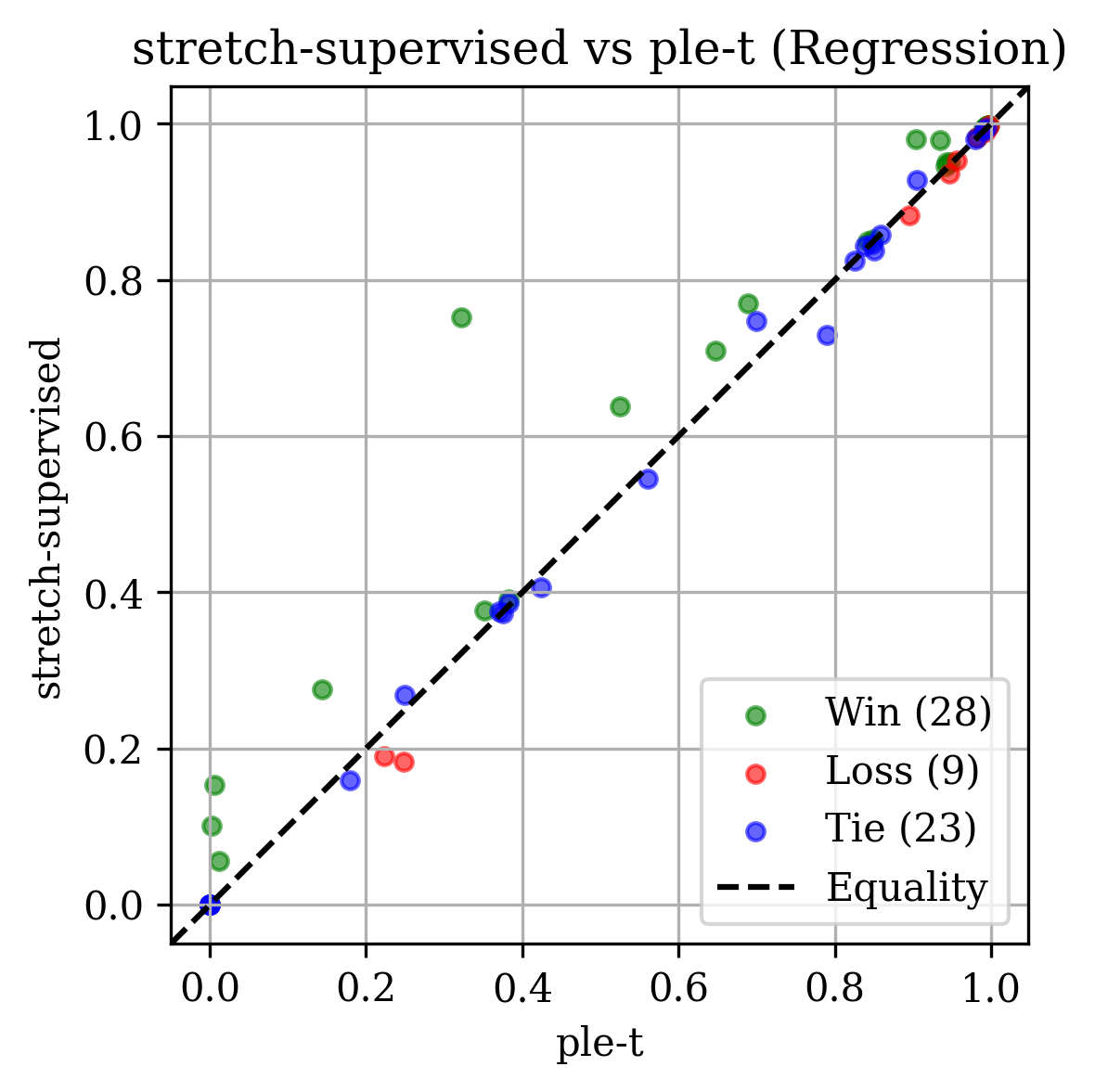}
    \caption{Stretch Supervised vs PLE-T}
    \label{fig:pairwise_scatter2}
\end{figure}

\begin{figure}[htbp]
    \centering
    \includegraphics[width=0.5\linewidth]{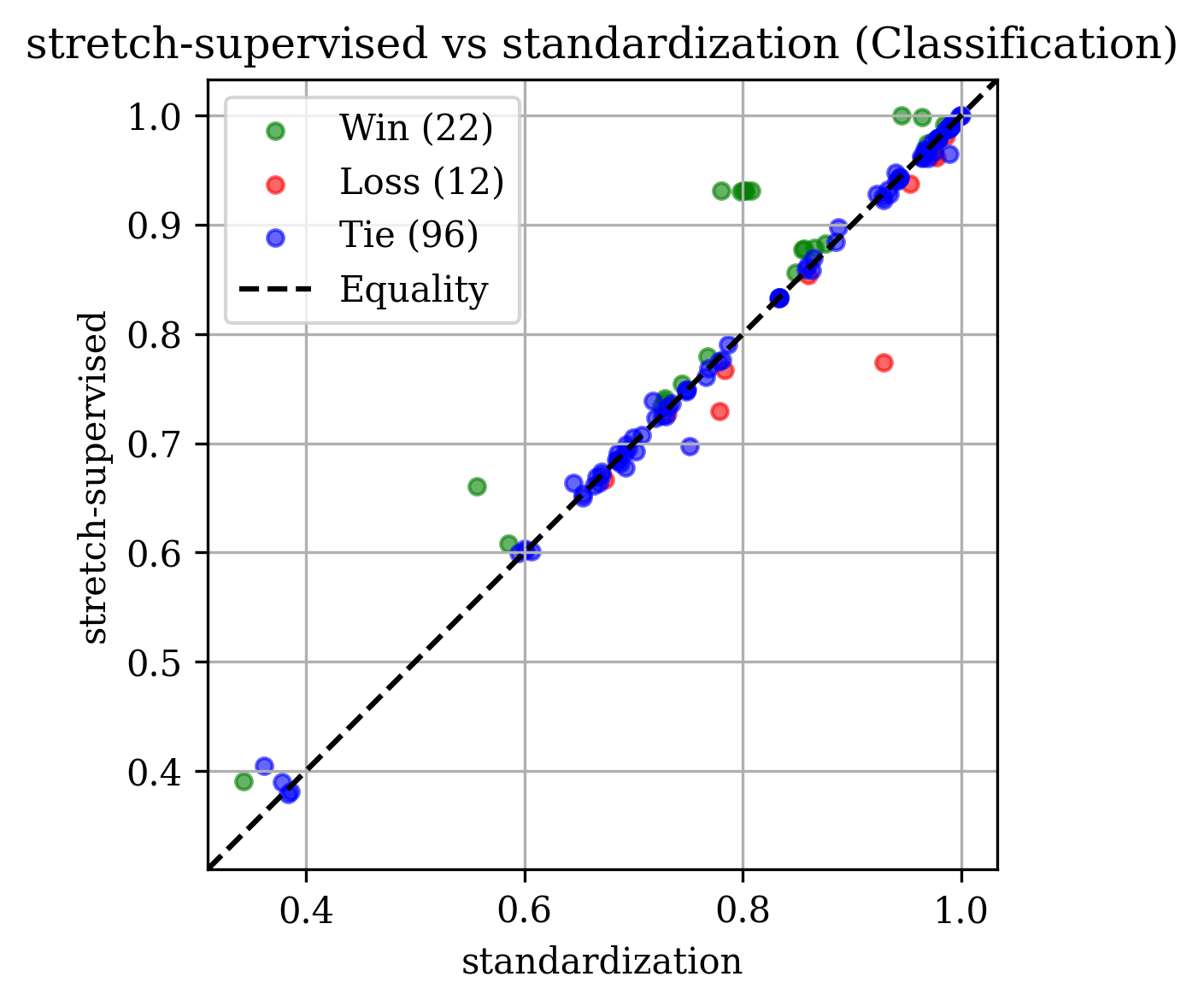}
    \hfill
    \includegraphics[width=0.49\linewidth]{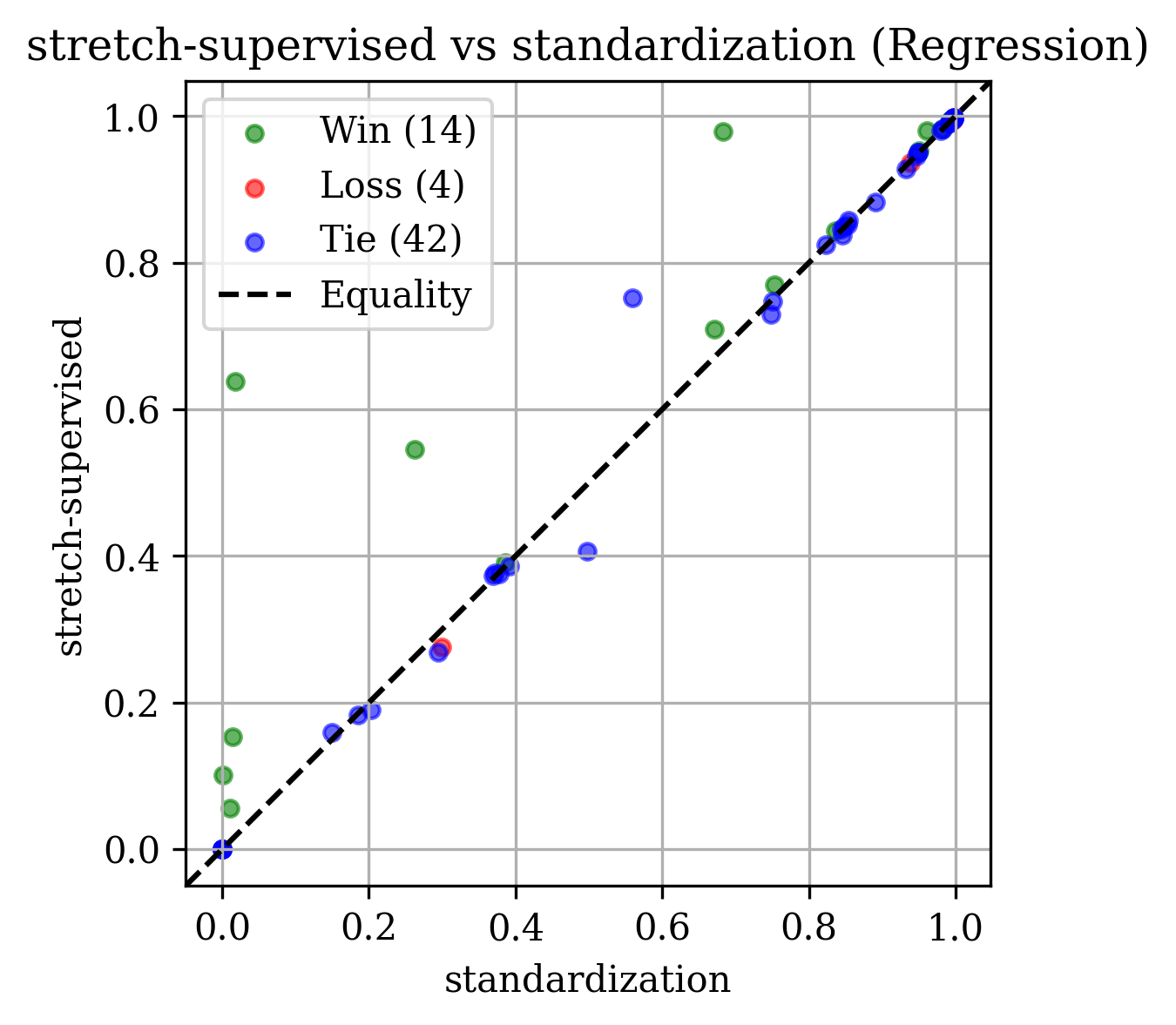}
    \caption{Stretch Supervised vs Standardization}
    \label{fig:pairwise_scatter3}
\end{figure}

\begin{figure}[htbp]
    \centering
    \includegraphics[width=0.455\linewidth]{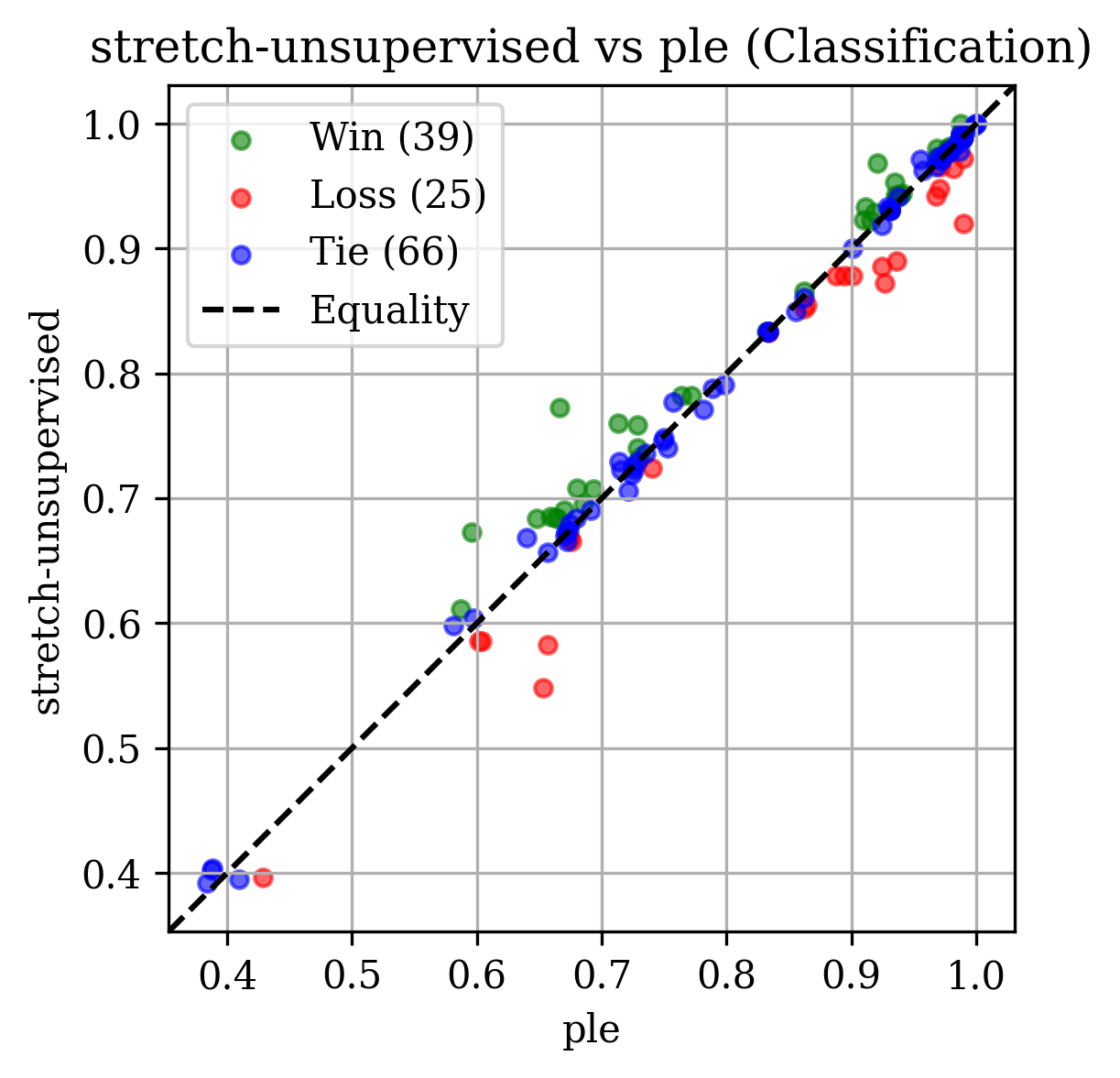}
    \hfill
    \includegraphics[width=0.47\linewidth]{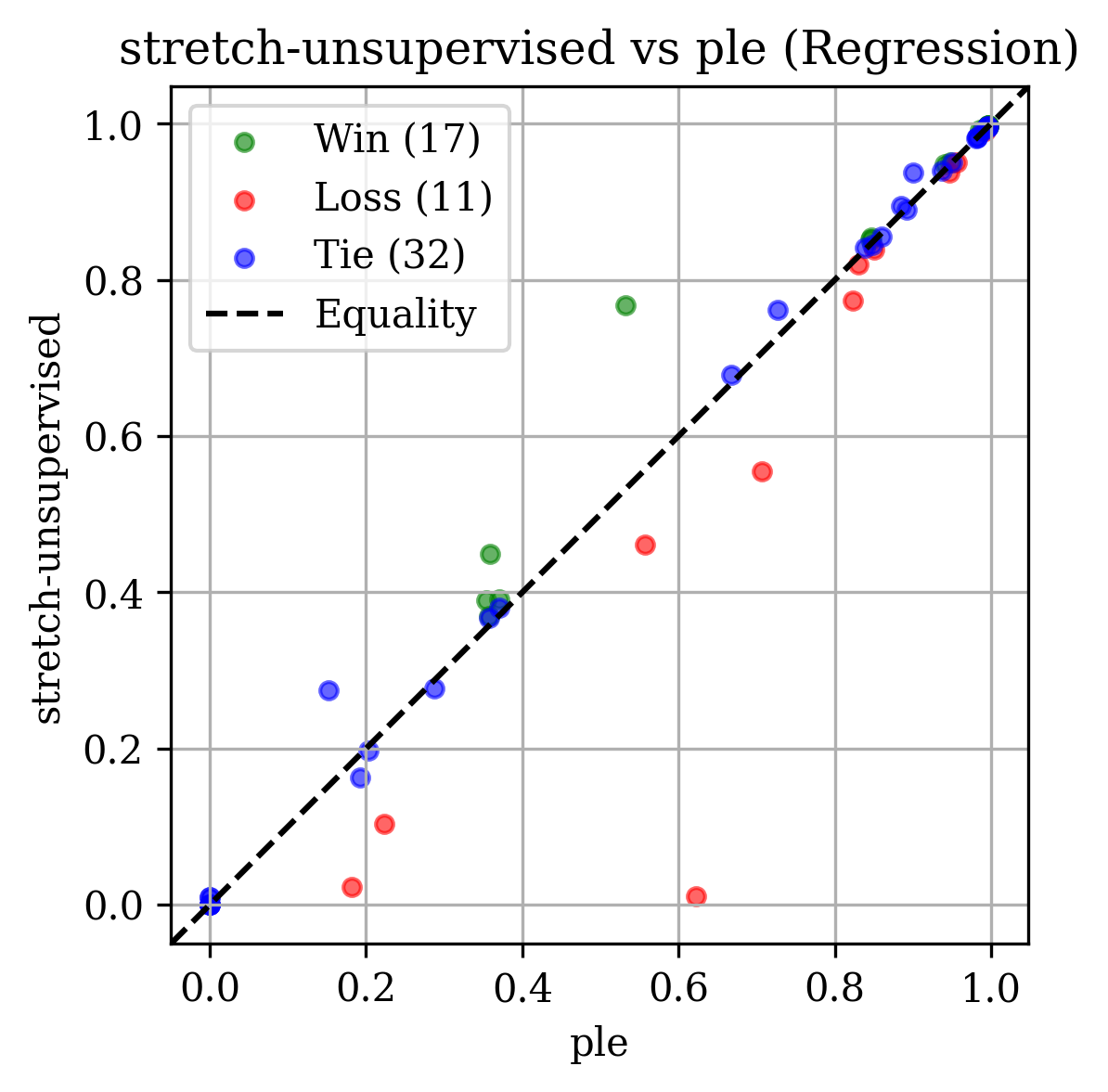}
    \caption{Stretch Unsupervised vs PLE}
    \label{fig:pairwise_scatter4}
\end{figure}

\begin{figure}[htbp]
    \centering
    \includegraphics[width=0.46\linewidth]{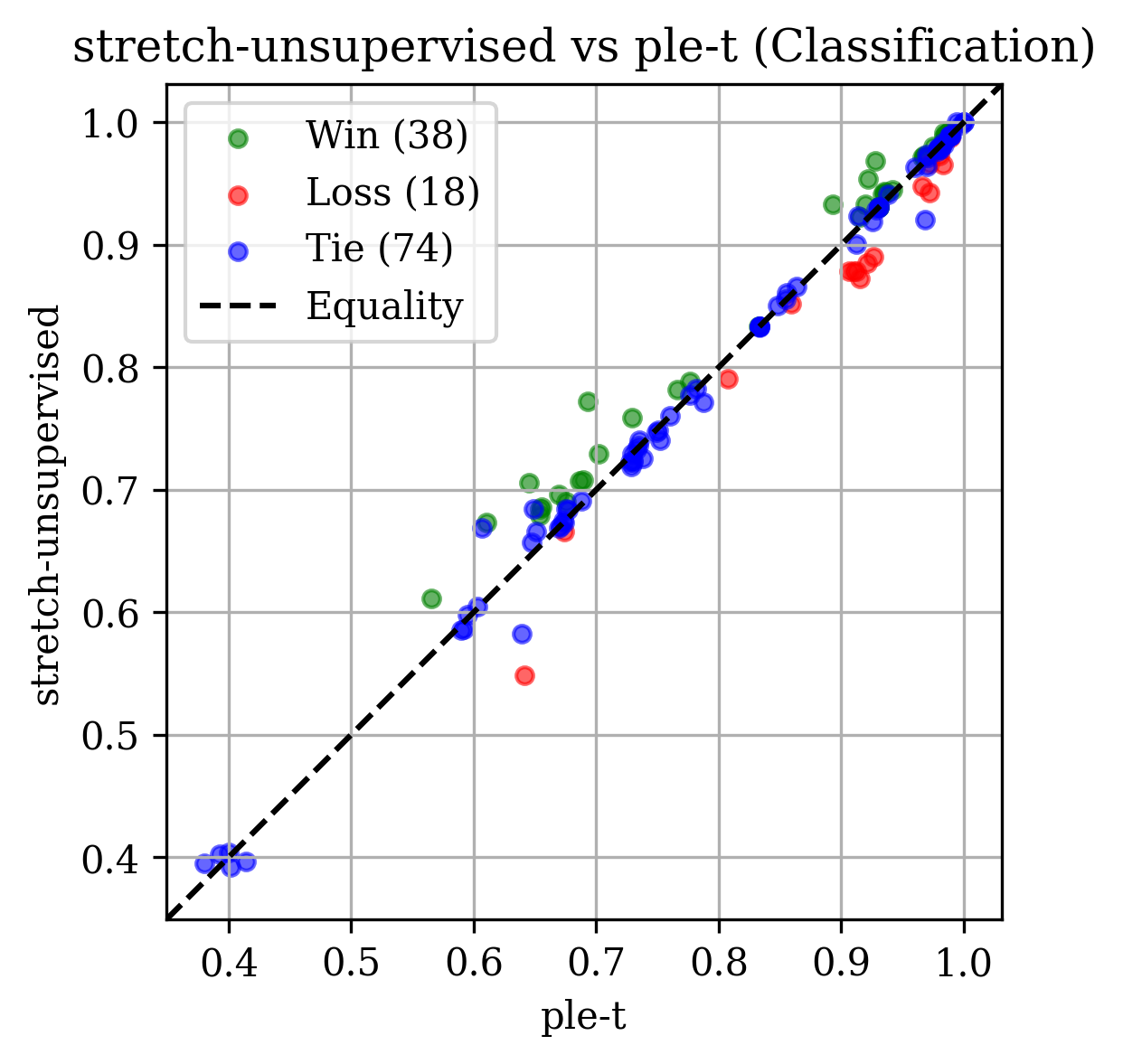}
    \hfill
    \includegraphics[width=0.46\linewidth]{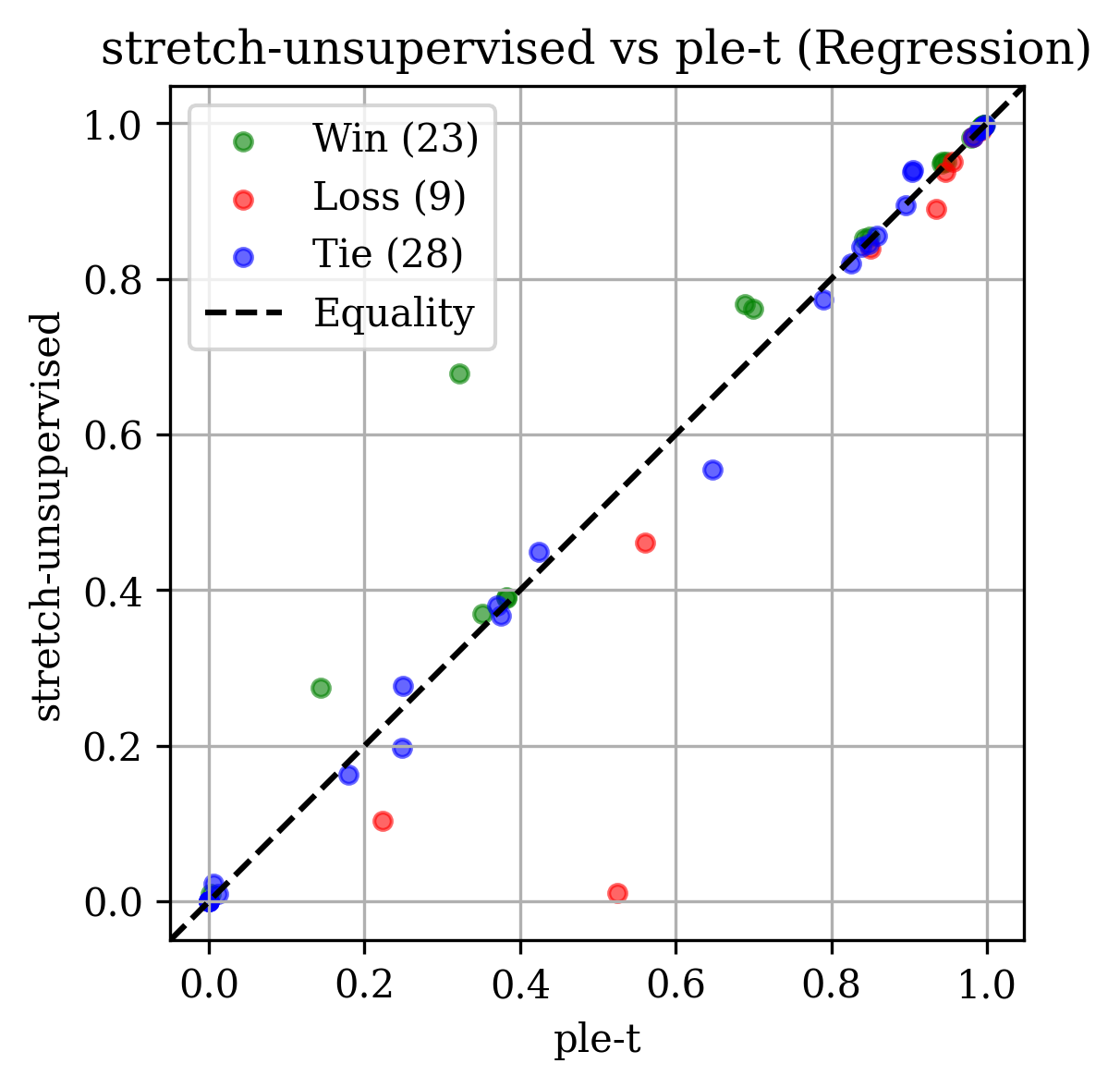}
    \caption{Stretch Unsupervised vs PLE-T}
    \label{fig:pairwise_scatter5}
\end{figure}

\begin{figure}[htbp]
    \centering
    \includegraphics[width=0.49\linewidth]{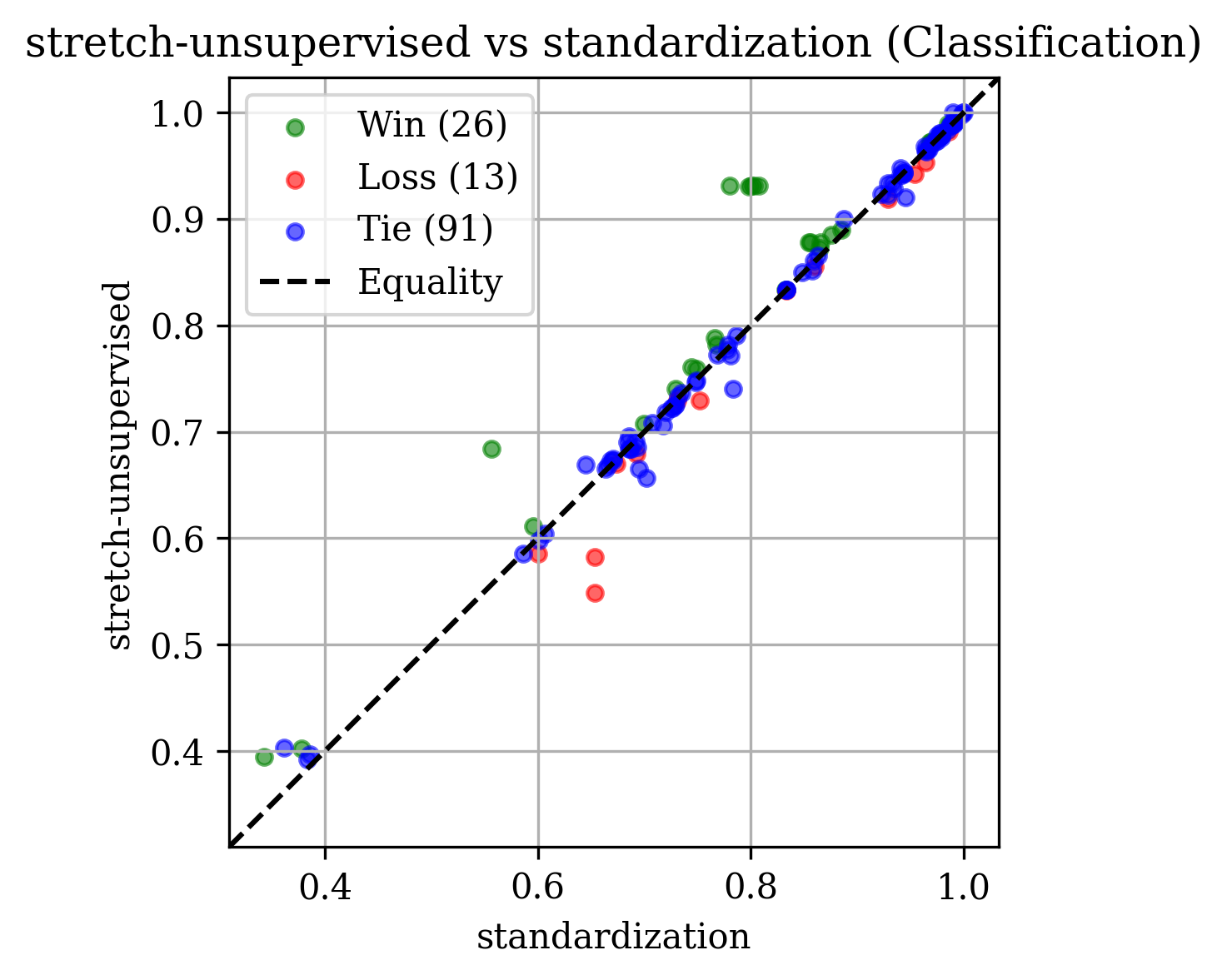}
    \hfill
    \includegraphics[width=0.49\linewidth]{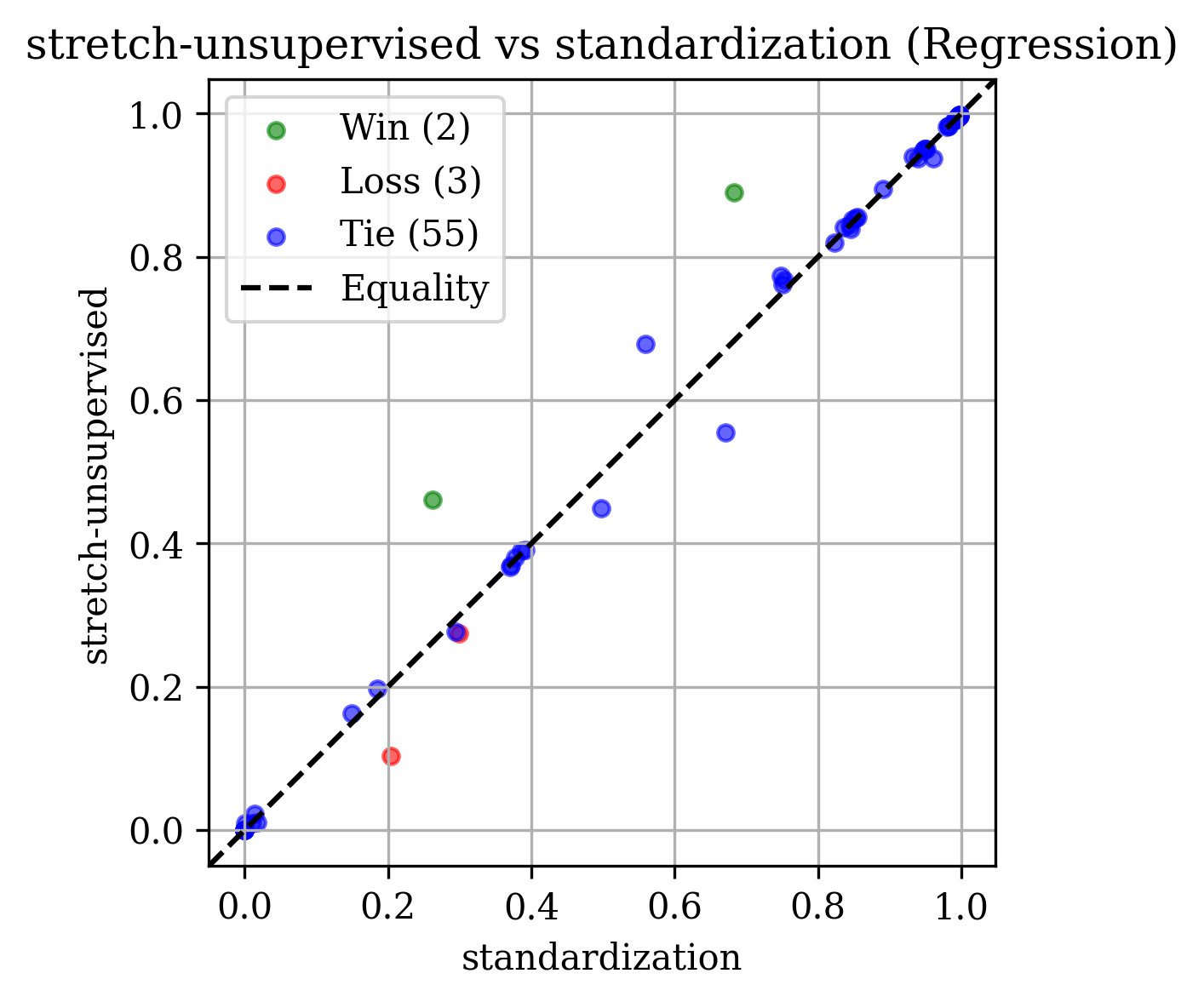}
    \caption{Stretch Unsupervised vs Standardization}
    \label{fig:pairwise_scatter6}
\end{figure}

Here, we look closer at these raw counts. Figures~\ref{fig:pairwise_scatter1} through \ref{fig:pairwise_scatter6} present the detailed scatter plots comparing our supervised and unsupervised methods against three key baselines: PLE, PLE-T, and standardization. In each figure, every point represents a single dataset-model combination. Points located above the diagonal indicate a win for the stretch transformation, points below indicate a loss, and points on the diagonal represent a statistical tie. The total win-loss-tie counts are summarized and displayed directly on each plot, providing the quantitative confirmation for the performance advantages discussed in Section~\ref{sec:analysis}.

\clearpage
\section{Dataset Composition and Feature Types}
\label{app:datasetcomposition}

The 38 datasets in our benchmark span a wide range of feature compositions. To make this explicit, we group them by the relative proportion of numeric and categorical features:
\begin{itemize}[leftmargin=1.5em, topsep=0pt, itemsep=0pt]
    \item \textbf{Numeric Dominant} (26 datasets): numeric features clearly outnumber categorical ones, including datasets that are purely numeric.
    \item \textbf{Balanced} (8 datasets): the two feature types are comparable in number (within a 2:1 ratio in either direction).
    \item \textbf{Category Dominant} (4 datasets): categorical features dominate, with at least a 3:1 ratio over numeric features.
\end{itemize}

Table~\ref{tab:dataset_characteristics} lists the per-dataset statistics. Sample sizes range from about 650 to over 65,000, and feature counts from a handful up to 140. The imbalance ratio column reports class imbalance for classification tasks where applicable.

Since stretch transformations only act on numeric features, the relative amount of numeric information differs substantially across these three groups. We use this grouping in Appendix~\ref{sec:performance_breakdown} (Table~\ref{tab:category_score_performance}) to check whether the gains from stretch persist as the share of numeric features decreases.

\begin{table}[t]
\centering
\caption{Dataset characteristics grouped by feature type dominance.}
\label{tab:dataset_characteristics}
\begin{adjustbox}{max width=\textwidth}
\begin{tabular}{l c c c c c}
\toprule
\textbf{Dataset} & \textbf{Num Features} & \textbf{Cat Features} & \textbf{Total Samples} & \textbf{Imbalance Ratio} & \textbf{N Classes} \\
\midrule
\multicolumn{6}{c}{\textbf{Balanced Datasets}} \\
\midrule
Diamonds & 6 & 3 & 53940 & -- & -- \\
E-CommereShippingData & 6 & 4 & 10999 & -- & -- \\
Fitness\_Club\_c & 3 & 3 & 1500 & -- & -- \\
Kaggle\_bike\_sharing\_demand & 3 & 6 & 10886 & -- & 1 \\
NHANES\_age\_prediction & 4 & 3 & 2277 & -- & -- \\
Shop\_Customer\_Data & 4 & 2 & 2000 & -- & -- \\
compass & 8 & 9 & 16644 & 1.000 & -- \\
estimation\_of\_obesity\_levels & 8 & 8 & 2111 & -- & -- \\
\midrule
\multicolumn{6}{c}{\textbf{Category Dominant Datasets}} \\
\midrule
archive\_r56\_Portuguese & 1 & 29 & 651 & -- & -- \\
okcupid\_stem & 2 & 11 & 26677 & 6.826 & 3 \\
socmob & 1 & 4 & 1156 & -- & -- \\
thyroid-dis & 6 & 20 & 2800 & 52.645 & 5 \\
\midrule
\multicolumn{6}{c}{\textbf{Numeric Dominant Datasets}} \\
\midrule
ASP-POTASSCO-classification & 140 & 1 & 1294 & 12.286 & 11 \\
Ailerons & 40 & 0 & 13750 & -- & -- \\
Bias\_correction\_r\_2 & 21 & 0 & 7725 & -- & -- \\
Click\_prediction\_small & 3 & 0 & 39948 & -- & -- \\
Contaminant-detection-11.0GHz & 30 & 0 & 2400 & -- & -- \\
FOREX\_audjpy-day-High & 10 & 0 & 1832 & 1.056 & 2 \\
FOREX\_audsgd-hour-High & 10 & 0 & 43825 & 1.061 & 2 \\
IEEE80211aa-GATS & 27 & 0 & 4046 & -- & -- \\
Intersectional-Bias-Assessment & 14 & 5 & 11000 & -- & 2 \\
Job\_Profitability & 27 & 1 & 14480 & -- & -- \\
Mobile\_Price\_Classification & 14 & 6 & 2000 & -- & -- \\
VulNoneVul & 16 & 0 & 5692 & -- & 2 \\
Waterstress & 22 & 0 & 1188 & -- & 2 \\
electricity & 7 & 1 & 45312 & 1.355 & 2 \\
htru & 8 & 0 & 17898 & -- & -- \\
ibm-employee-performance & 23 & 7 & 1470 & 5.504 & 2 \\
internet\_firewall & 7 & 0 & 65532 & -- & -- \\
jungle\_chess\_2pcs\_endgame & 6 & 0 & 44819 & 5.320 & 3 \\
kc1 & 21 & 0 & 2109 & 5.469 & 2 \\
optdigits & 64 & 0 & 5620 & 1.032 & 10 \\
page-blocks & 10 & 0 & 5473 & 175.464 & 5 \\
pol & 26 & 0 & 10082 & 1.000 & 2 \\
pol\_reg & 48 & 0 & 15000 & -- & -- \\
pole & 26 & 0 & 14998 & -- & -- \\
wine & 4 & 0 & 2554 & 1.000 & 2 \\
yeast & 8 & 0 & 1484 & 92.600 & 10 \\
\bottomrule
\end{tabular}
\end{adjustbox}
\end{table}

\clearpage
\section{Performance Breakdown by Feature Composition} \label{sec:performance_breakdown}
Table~\ref{tab:category_score_performance} reports per-model average scores within each of the three dataset groups defined in Appendix~\ref{app:datasetcomposition}. The main observations are:
\begin{itemize}[leftmargin=1.5em, topsep=0pt, itemsep=0pt]
\item \textbf{Numeric Dominant}: Supervised Stretch is the top method on most models, which is the regime where our objective is most directly applicable.
\item \textbf{Balanced}: results are more mixed across methods, with Supervised Stretch and PLE-T trading wins on different model-task pairs, and no single method dominating.
\item \textbf{Category Dominant}: Despite numeric features being a minority, the stretch variants remain competitive. Supervised Stretch attains the highest regression Overall Score, and Unsupervised Stretch the highest classification Overall Score in this group.
\end{itemize}

\begin{table}[h]
\centering
\caption{Performance by dataset category and task type (average score).}
\label{tab:category_score_performance}
\begin{adjustbox}{max width=\textwidth}
\begin{tabular}{l l l ccccccccc}
\toprule
\textbf{Category} & \textbf{Task} & \textbf{Model} & \textbf{Sup.} & \textbf{Unsup.} & \textbf{Minmax} & \textbf{Ple} & \textbf{Ple T} & \textbf{Quantile} & \textbf{RS-SC} & \textbf{Std} & \textbf{YJ} \\
\midrule
\multirow{12}{*}{Balanced}
    & \multirow{6}{*}{Classification} & ftt & 0.7987 & 0.8000 & 0.7911 & 0.7967 & 0.7994 & \textbf{0.8023} & 0.8005 & 0.7987 & 0.8014 \\
    & & mlp & 0.7673 & 0.7827 & 0.7738 & 0.7860 & \textbf{0.7907} & 0.7809 & 0.7855 & 0.7813 & 0.7849 \\
    & & mlp\_plr & 0.7843 & \textbf{0.7994} & 0.7850 & 0.7891 & 0.7880 & 0.7875 & 0.7926 & 0.7916 & 0.7932 \\
    & & realmlp & 0.8008 & 0.8020 & 0.7984 & 0.7990 & \textbf{0.8101} & 0.8021 & 0.8001 & 0.8005 & 0.8009 \\
    & & resnet & 0.7844 & 0.7793 & 0.7735 & 0.7777 & \textbf{0.7871} & 0.7795 & 0.7817 & 0.7834 & 0.7868 \\
    & & \textbf{Overall} & 0.7871 & 0.7927 & 0.7844 & 0.7897 & \textbf{0.7950} & 0.7905 & 0.7921 & 0.7911 & 0.7934 \\
    \cmidrule(lr){2-12}
    & \multirow{6}{*}{Regression} & ftt & 0.5493 & 0.5478 & 0.5422 & 0.5418 & 0.5477 & 0.5482 & \textbf{0.5503} & 0.5479 & \textbf{0.5503} \\
    & & mlp & 0.4404 & 0.4300 & 0.4740 & 0.4051 & 0.4321 & 0.4747 & 0.4736 & 0.3896 & \textbf{0.4961} \\
    & & mlp\_plr & 0.5340 & 0.5346 & \textbf{0.5375} & 0.4708 & 0.5128 & 0.5313 & 0.5346 & 0.5311 & 0.5327 \\
    & & realmlp & \textbf{0.5501} & 0.5475 & 0.5357 & 0.5445 & 0.5491 & 0.5419 & 0.5478 & 0.5471 & 0.5435 \\
    & & resnet & 0.5259 & 0.5172 & 0.5225 & 0.4962 & 0.4889 & 0.5274 & \textbf{0.5326} & 0.5207 & 0.5254 \\
    & & \textbf{Overall} & 0.5199 & 0.5154 & 0.5224 & 0.4917 & 0.5061 & 0.5247 & 0.5278 & 0.5073 & \textbf{0.5296} \\
\midrule
\multirow{12}{*}{Category Dominant}
    & \multirow{6}{*}{Classification} & ftt & \textbf{0.7207} & 0.7196 & 0.7185 & 0.7206 & 0.7191 & 0.7161 & 0.7199 & 0.7205 & 0.7119 \\
    & & mlp & 0.7063 & \textbf{0.7125} & 0.7054 & 0.7072 & 0.6993 & 0.7030 & 0.7102 & 0.7085 & 0.7077 \\
    & & mlp\_plr & 0.7076 & 0.7086 & 0.7093 & 0.6978 & 0.7046 & 0.7082 & \textbf{0.7128} & 0.7098 & 0.7093 \\
    & & realmlp & \textbf{0.7202} & 0.7154 & 0.7168 & 0.7141 & 0.7122 & 0.7160 & 0.7173 & 0.7169 & 0.7160 \\
    & & resnet & 0.7107 & 0.7133 & \textbf{0.7171} & 0.7025 & 0.7049 & 0.7091 & 0.7077 & 0.7094 & 0.7116 \\
    & & \textbf{Overall} & 0.7131 & \textbf{0.7139} & 0.7134 & 0.7084 & 0.7080 & 0.7105 & 0.7136 & 0.7130 & 0.7113 \\
    \cmidrule(lr){2-12}
    & \multirow{6}{*}{Regression} & ftt & 0.5797 & 0.5845 & 0.5241 & 0.5178 & 0.5191 & 0.5790 & 0.5374 & \textbf{0.5951} & 0.5857 \\
    & & mlp & \textbf{0.4713} & 0.3908 & 0.4261 & 0.4453 & 0.2726 & 0.4377 & 0.3772 & 0.3813 & 0.4507 \\
    & & mlp\_plr & 0.4564 & 0.4850 & 0.4760 & 0.5130 & \textbf{0.5190} & 0.4576 & 0.4366 & 0.4666 & 0.4646 \\
    & & realmlp & 0.5984 & 0.6088 & 0.5979 & 0.6122 & 0.5774 & \textbf{0.6251} & 0.5889 & 0.6134 & 0.5854 \\
    & & resnet & 0.4342 & 0.3587 & \textbf{0.4895} & 0.4493 & 0.4134 & 0.3550 & 0.3818 & 0.4104 & 0.3532 \\
    & & \textbf{Overall} & \textbf{0.5080} & 0.4856 & 0.5027 & 0.5076 & 0.4603 & 0.4909 & 0.4644 & 0.4934 & 0.4879 \\
\midrule
\multirow{12}{*}{Numeric Dominant}
    & \multirow{6}{*}{Classification} & ftt & \textbf{0.8340} & 0.8287 & 0.8048 & 0.8315 & 0.8305 & 0.8331 & 0.8216 & 0.8179 & 0.8252 \\
    & & mlp & 0.8313 & 0.8326 & 0.8234 & \textbf{0.8362} & 0.8283 & 0.8270 & 0.8306 & 0.8254 & 0.8294 \\
    & & mlp\_plr & 0.8331 & 0.8309 & 0.8269 & 0.8279 & 0.8273 & \textbf{0.8348} & 0.8278 & 0.8250 & 0.8255 \\
    & & realmlp & 0.8444 & 0.8451 & \textbf{0.8477} & 0.8401 & 0.8392 & 0.8450 & 0.8459 & 0.8379 & 0.8401 \\
    & & resnet & \textbf{0.8394} & 0.8374 & 0.8248 & 0.8366 & 0.8339 & 0.8296 & 0.8387 & 0.8301 & 0.8307 \\
    & & \textbf{Overall} & \textbf{0.8364} & 0.8349 & 0.8255 & 0.8345 & 0.8318 & 0.8339 & 0.8329 & 0.8273 & 0.8302 \\
    \cmidrule(lr){2-12}
    & \multirow{6}{*}{Regression} & ftt & 0.8228 & 0.8017 & 0.7964 & \textbf{0.8260} & 0.7961 & 0.8041 & 0.7991 & 0.7993 & 0.7716 \\
    & & mlp & \textbf{0.8100} & 0.7951 & 0.7977 & 0.7969 & 0.7964 & 0.7757 & 0.7969 & 0.7950 & 0.6096 \\
    & & mlp\_plr & 0.8884 & 0.8746 & 0.8011 & 0.8868 & \textbf{0.8892} & 0.8310 & 0.8600 & 0.8413 & 0.7879 \\
    & & realmlp & \textbf{0.9063} & 0.8013 & 0.8016 & 0.9043 & 0.8884 & 0.8435 & 0.8162 & 0.8023 & 0.8089 \\
    & & resnet & \textbf{0.8060} & 0.7980 & 0.7979 & 0.7966 & 0.7977 & 0.7808 & 0.7974 & 0.7979 & 0.6162 \\
    & & \textbf{Overall} & \textbf{0.8467} & 0.8141 & 0.7990 & 0.8421 & 0.8335 & 0.8070 & 0.8139 & 0.8072 & 0.7188 \\
\bottomrule
\end{tabular}
\end{adjustbox}
\end{table}

\clearpage
\section{Marginal Signal Non-Uniformity Analysis}
\label{app:marginal_snalysis}

Supervised stretch is most useful when a numeric feature's marginal effect on the target is non-uniform, i.e., when there are local regions where the target changes much faster than elsewhere. A natural concern is whether this is actually a common pattern in tabular data, or whether most features have roughly flat marginal effects on which our method would have little to do. This appendix measures the degree of non-uniformity directly on the benchmark datasets.

\subsection{Methodology}

For each numeric feature in each benchmark dataset, we proceed as follows:
\begin{enumerate}[leftmargin=1.5em, topsep=0pt, itemsep=0pt]
    \item \textbf{Preprocessing.} Mean-impute missing values and standardize to zero mean and unit variance.
    \item \textbf{Marginal function.} For each unique feature value $x$, compute $f(x) = \mathbb{E}[t \mid x]$. For regression, this is the mean of the standardized target; for classification, the empirical class-probability vector $f(x) \in \mathbb{R}^K$ (see footnote~\ref{fn:target}).
    \item \textbf{Local slopes.} For consecutive unique values $x_i, x_{i+1}$, compute $|\Delta f_i / \Delta x_i|$, where $|\cdot|$ denotes absolute value (regression) or $\ell_2$ norm (classification).
    \item \textbf{Normalization.} Divide each local slope by the per-feature mean so that the rescaled slope is
    \begin{equation}
        r_i = \frac{|\Delta f_i / \Delta x_i| \cdot n}{\sum_{j=1}^{n} |\Delta f_j / \Delta x_j|},
    \end{equation}
    where $n$ is the number of local slopes for that feature. This makes each feature's slopes have mean $1$, so we are looking at the \emph{shape} of the slope distribution, not its overall magnitude. A perfectly uniform marginal effect gives $r_i \equiv 1$; non-uniformity shows up as values both above and below $1$.
    \item \textbf{Aggregation.} Pool the $r_i$ values from all numeric features in a dataset into one distribution.
    \item \textbf{Summary statistics.} Compute the standard deviation and skewness of this pooled distribution per dataset. Standard deviation captures how spread out the slopes are; positive skewness flags the presence of a few regions with much larger slopes than the rest.
\end{enumerate}

\subsection{Results}

Table~\ref{tab:marginal_std_skewness_final} reports the standard deviation and skewness for every dataset in the benchmark, together with four image datasets included as a reference point.

\textbf{Tabular features are clearly non-uniform.} Standard deviations on the tabular datasets range from $0.47$ to $20.15$ with a median around $1.1$. If marginal effects were close to flat, the rescaled slopes would all be near $1$ and the standard deviations would be small. Instead, the values are large enough that, on most features, some regions have local slopes several times larger than others. Skewness is positive on most datasets, which means the distribution has a heavy right tail: a small number of regions are much steeper than the rest of the feature.

\textbf{Image data behaves differently.} For comparison, MNIST \citep{mnist}, Fashion-MNIST \citep{fashion_mnist}, CIFAR-10 and CIFAR-100 \citep{cifar} all sit between $0.28$ and $0.39$ in standard deviation, and have small to moderate skewness. Image features are pixel intensities on a fixed grid, so neighbouring pixels are equally spaced and bounded by the same range; the marginal slope distribution is correspondingly more uniform. Numeric tabular features have no such grid: samples can be arbitrarily close in feature space and targets in regression are unbounded, which is consistent with the much larger spread we see.

\textbf{Non-uniformity is not specific to numeric-dominant datasets.} The same pattern shows up across all three groups of Appendix~\ref{app:datasetcomposition}. Even category-dominant datasets contain numeric features with substantial non-uniformity, e.g., \texttt{socmob} (std $=2.03$, skewness $=6.26$) and \texttt{thyroid-dis} (std $=0.94$, skewness $=1.20$). What matters is the shape of the slope distribution per numeric feature, not how many numeric features the dataset has overall.

\textbf{Connection to method behaviour.} Supervised stretch reacts to exactly the kind of structure measured here. In bins where the target varies rapidly the bin widths grow; in bins where it varies little they shrink. When a feature actually is close to uniform, the bin-wise total variations $S_t$ are similar across bins and the allocation collapses back towards uniform widths, so the transformation has little effect. The fact that most features in the benchmark have non-trivial dispersion and skewness suggests there is in practice plenty of structure for the method to act on, which is consistent with the empirical gains in Section~\ref{sec:experiments}.

\begin{table}[t]
\centering
\caption{Marginal distribution metrics (Std and Skewness)}
\label{tab:marginal_std_skewness_final}
\small
\begin{adjustbox}{max width=\textwidth}
\begin{tabular}{l c c c}
\toprule
\textbf{Dataset} & \textbf{Task Type} & \textbf{Std} & \textbf{Skewness} \\
\midrule
\multicolumn{4}{c}{\textbf{Balanced Datasets}} \\
\midrule
Diamonds & Regression & 2.1275 & 5.52 \\
E-CommereShippingData & Binary & 1.3061 & 10.28 \\
Fitness\_Club\_c & Binary & 1.2097 & 0.60 \\
Kaggle\_bike\_sharing\_demand\_challange & Regression & 1.0102 & 2.47 \\
NHANES\_age\_prediction & Regression & 0.9773 & 1.57 \\
Shop\_Customer\_Data & Regression & 2.6387 & 6.50 \\
compass & Binary & 0.9258 & 1.02 \\
estimation\_of\_obesity\_levels & Multiclass & 0.7353 & -0.05 \\
\midrule
\multicolumn{4}{c}{\textbf{Category Dominant Datasets}} \\
\midrule
archive\_r56\_Portuguese & Regression & 1.1955 & 1.33 \\
okcupid\_stem & Multiclass & 1.1132 & 1.44 \\
socmob & Regression & 2.0314 & 6.26 \\
thyroid-dis & Multiclass & 0.9361 & 1.20 \\
\midrule
\multicolumn{4}{c}{\textbf{Numeric Dominant Datasets}} \\
\midrule
ASP-POTASSCO-classification & Multiclass & 0.4705 & -1.27 \\
Ailerons & Regression & 1.3136 & 3.48 \\
Bias\_correction\_r\_2 & Regression & 12.5727 & 53.98 \\
Click\_prediction\_small & Binary & 1.0574 & 0.72 \\
Contaminant-detection-11.0GHz & Binary & 1.0746 & 0.66 \\
FOREX\_audjpy-day-High & Binary & 1.0024 & 0.03 \\
FOREX\_audsgd-hour-High & Binary & 0.8862 & 0.09 \\
IEEE80211aa-GATS & Regression & 13.4710 & 61.56 \\
Intersectional-Bias-Assessment & Binary & 1.1410 & 0.33 \\
Job\_Profitability & Regression & 10.1227 & 73.20 \\
Mobile\_Price\_Classification & Multiclass & 0.8191 & 0.74 \\
VulNoneVul & Binary & 2.3204 & 3.53 \\
Waterstress & Binary & 1.0665 & 0.14 \\
electricity & Binary & 1.1502 & 1.65 \\
htru & Binary & 3.3842 & 4.15 \\
ibm-employee-performance & Binary & 1.5927 & 1.21 \\
internet\_firewall & Multiclass & 20.1475 & 53.98 \\
jungle\_chess\_2pcs\_endgame & Multiclass & 0.6012 & 0.50 \\
kc1 & Binary & 1.1874 & 0.64 \\
optdigits & Multiclass & 0.5452 & 2.06 \\
page-blocks & Multiclass & 2.0516 & 2.73 \\
pol & Binary & 1.6962 & 3.22 \\
pol\_reg & Regression & 1.6065 & 2.54 \\
pole & Regression & 1.5757 & 2.30 \\
wine & Binary & 1.0016 & 1.02 \\
yeast & Multiclass & 0.8551 & 1.20 \\
\midrule
\multicolumn{4}{c}{\textbf{Image Datasets}} \\
\midrule
MNIST & Multiclass & 0.353687 & 0.12 \\
Fashion-MNIST & Multiclass & 0.391164 & 0.94 \\
CIFAR-10 & Multiclass & 0.380173 & 1.47 \\
CIFAR-100 & Multiclass & 0.276569 & 1.75 \\
\bottomrule
\end{tabular}
\end{adjustbox}
\end{table}

\clearpage
\section{Noise Robustness Analysis}
\label{app:noise_analysis}

Since supervised stretch uses targets to design the transformation, a natural question is how it behaves when those targets are noisy. We test this by adding controlled label noise to four datasets and tracking how the relative ranking of methods changes with the noise level.

\subsection{Experimental Setup}

\textbf{Datasets and methods.} We use four datasets from our benchmark: \texttt{Contaminant-detection}, \texttt{yeast} and \texttt{ibm-employee-performance} (classification) and \texttt{NHANES-age-prediction} (regression). We compare Supervised Stretch and Unsupervised Stretch against Standardization, PLE, and PLE-T, on the same five architectures used in the main experiments.

\textbf{Noise injection.} For classification, we use symmetric label flipping: at noise level $\eta$, a fraction $\eta$ of training labels is replaced by a uniformly random label. For regression, we add Gaussian noise to training targets, $\epsilon \sim \mathcal{N}(0, \sigma_{\rm noise}^2)$ with $\sigma_{\rm noise} = \eta \cdot \sigma_{\rm target}$. Validation and test sets are kept clean, so the metric still measures recovery of the true signal.

\textbf{Evaluation.} Because accuracy and $R^2$ are not directly comparable across datasets, we report the \emph{mean rank} of each method. For every (dataset, model) pair and every noise level, the five methods are ranked from $1$ (best) to $5$ (worst). We then average these ranks over the $4 \times 5 = 20$ (dataset, model) combinations. Error bars in the figures show the standard deviation of the rank across these combinations.

\subsection{Results}

\begin{figure}[H]
    \centering
    \includegraphics[width=0.75\linewidth]{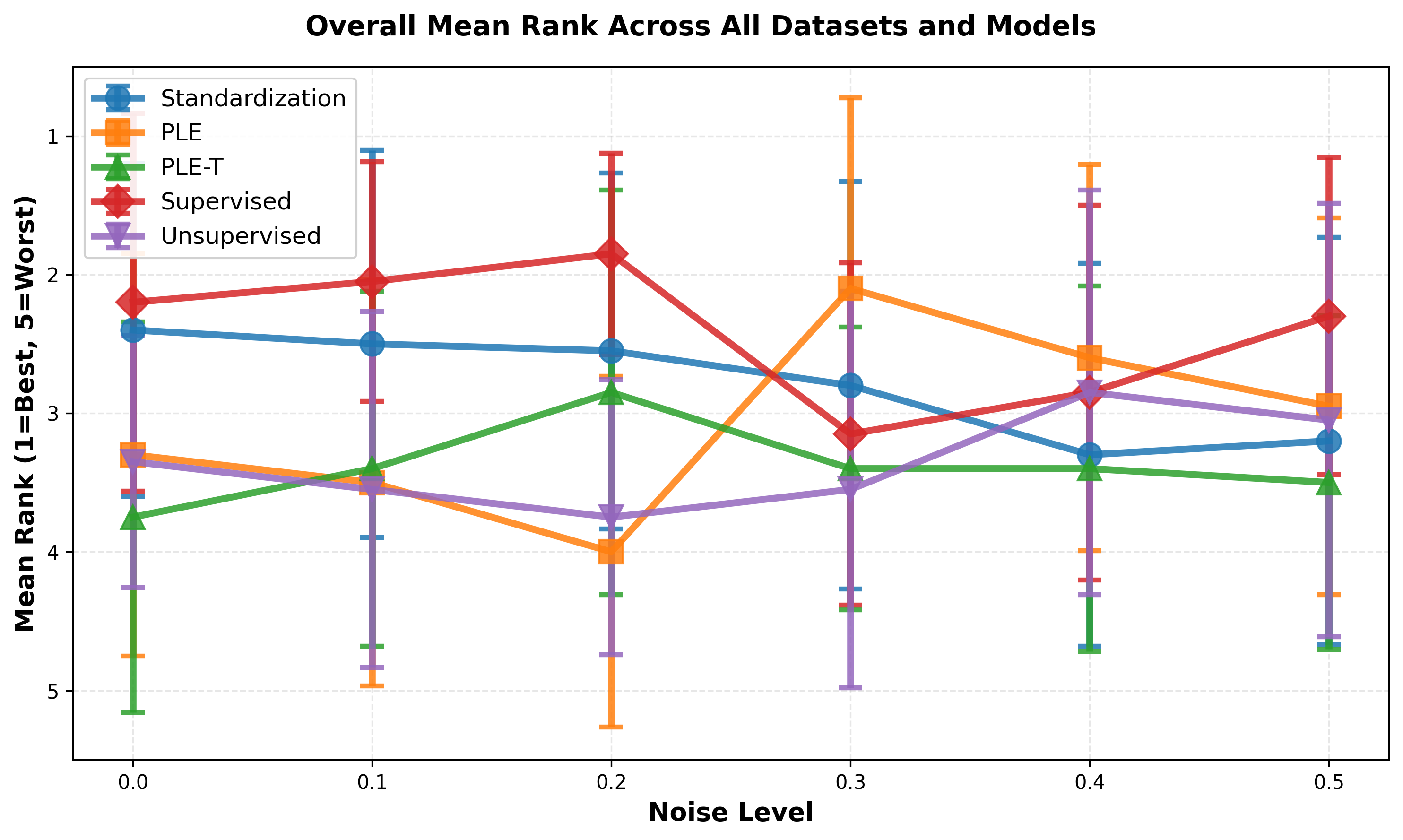}
    \caption{\textbf{Mean rank under label noise, aggregated over all (dataset, model) pairs.} Each method is ranked $1$--$5$ on every (dataset, model) pair at every noise level, then averaged over the $20$ combinations ($4$ datasets $\times\,5$ models). Lower is better. Supervised Stretch (red) attains the best mean rank at $\eta = 0.5$. Error bars show the standard deviation of ranks across (dataset, model) pairs.}
    \label{fig:noise_rank}
\end{figure}

\begin{figure}[H]
    \centering
    \begin{subfigure}[b]{0.48\linewidth}
        \centering
        \includegraphics[width=\linewidth]{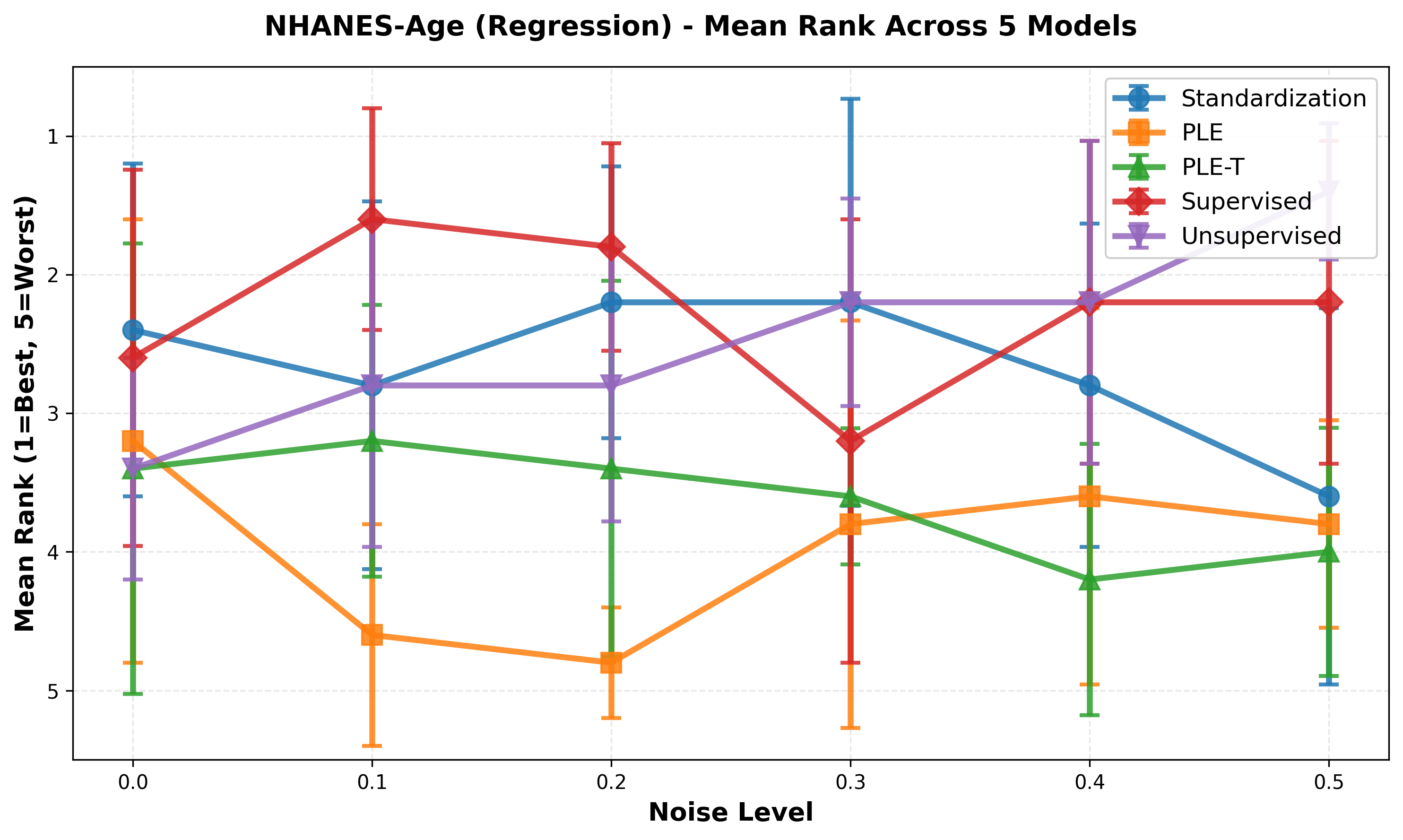}
        \caption{\texttt{NHANES-age-prediction} (Regression)}
        \label{fig:n1}
    \end{subfigure}
    \hfill
    \begin{subfigure}[b]{0.48\linewidth}
        \centering
        \includegraphics[width=\linewidth]{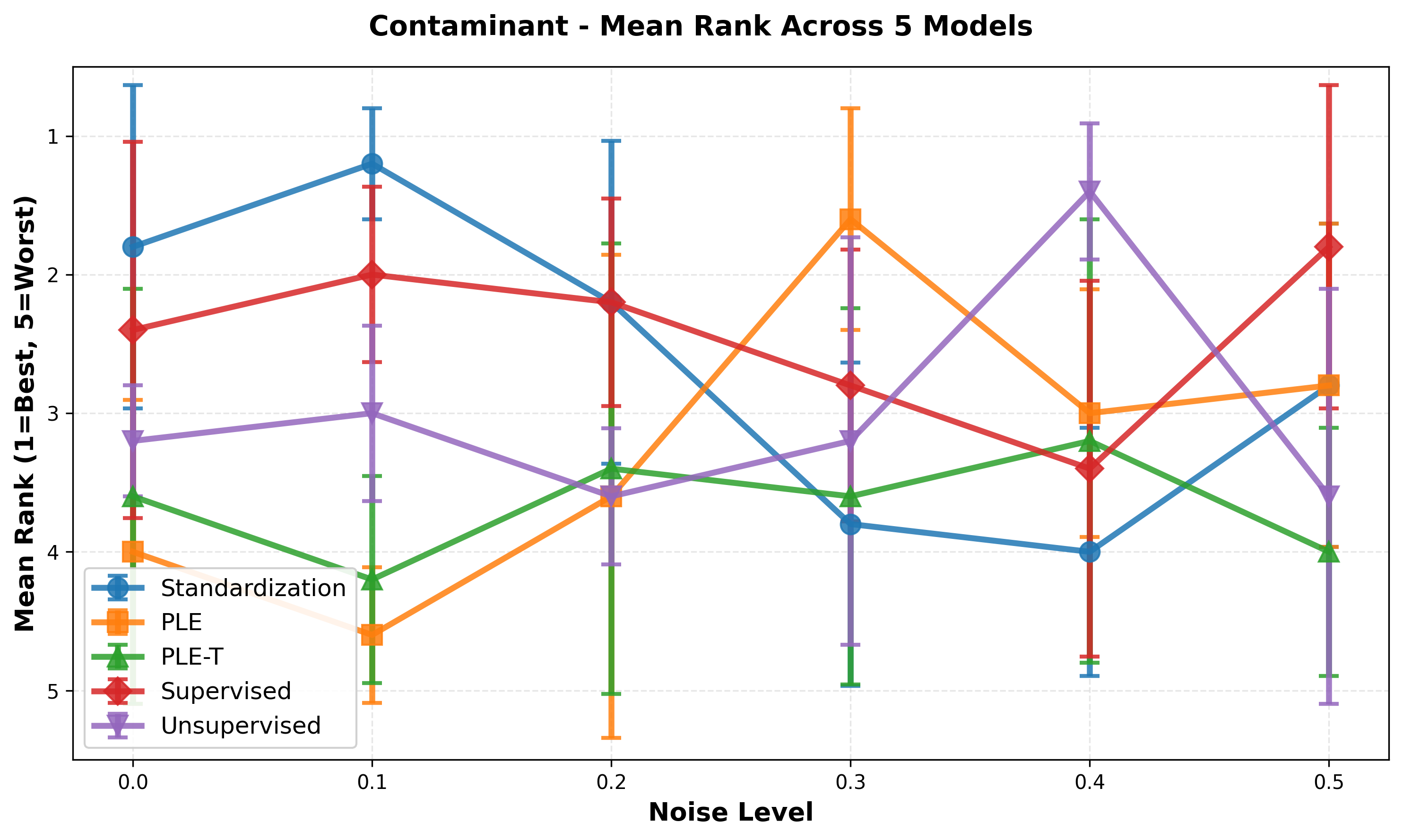}
        \caption{\texttt{Contaminant-detection} (Classification)}
        \label{fig:n2}
    \end{subfigure}
    \caption{\textbf{Detailed Mean Rank Trajectories (1/2).} Mean rank across 5 models for \texttt{NHANES-age} and \texttt{Contaminant} datasets under varying noise levels.}
    \label{fig:noise_rank_part1}
\end{figure}

\begin{figure}[H]
    \centering
    \begin{subfigure}[b]{0.48\linewidth}
        \centering
        \includegraphics[width=\linewidth]{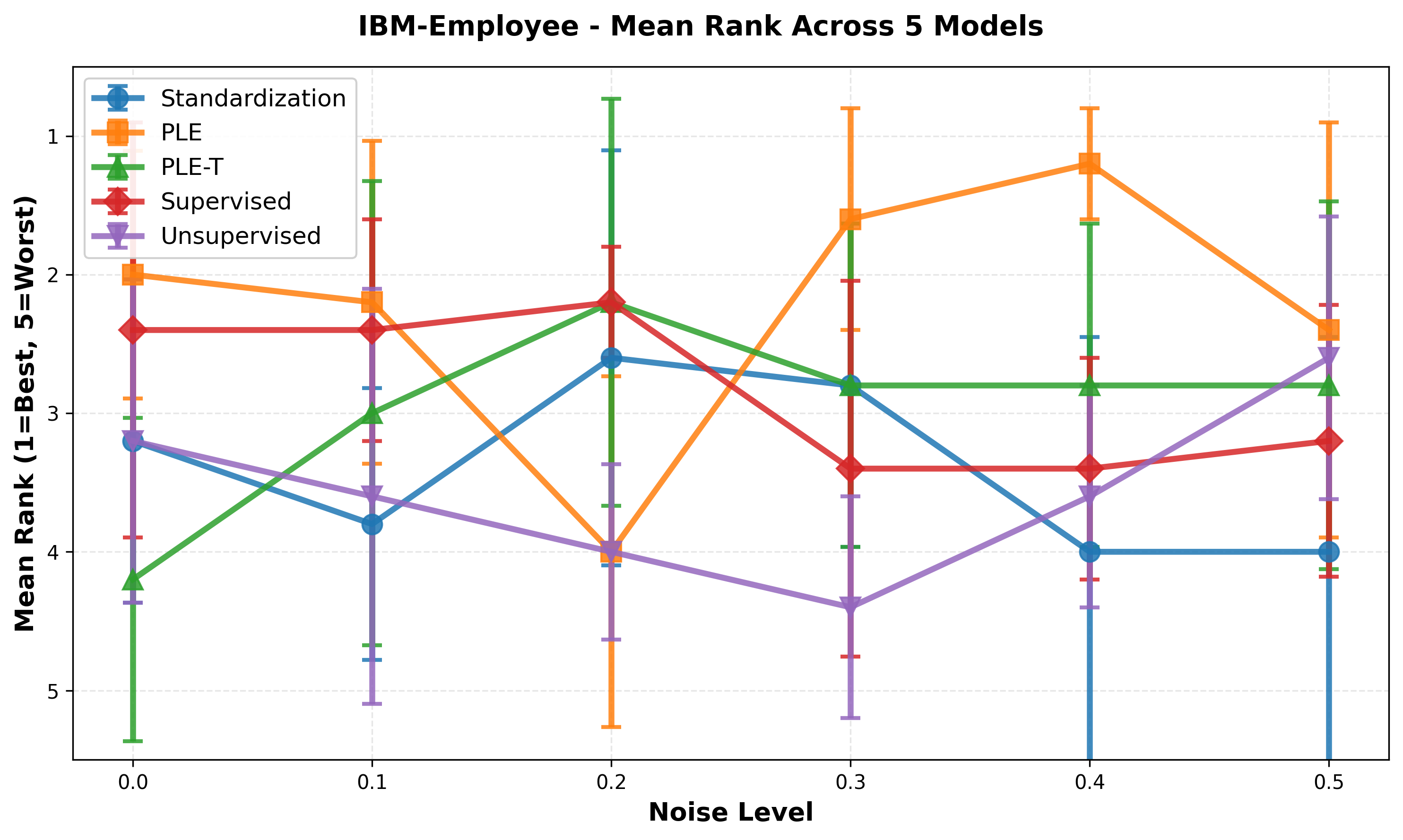}
        \caption{\texttt{ibm-employee-performance} (Classification)}
        \label{fig:n3}
    \end{subfigure}
    \hfill
    \begin{subfigure}[b]{0.48\linewidth}
        \centering
        \includegraphics[width=\linewidth]{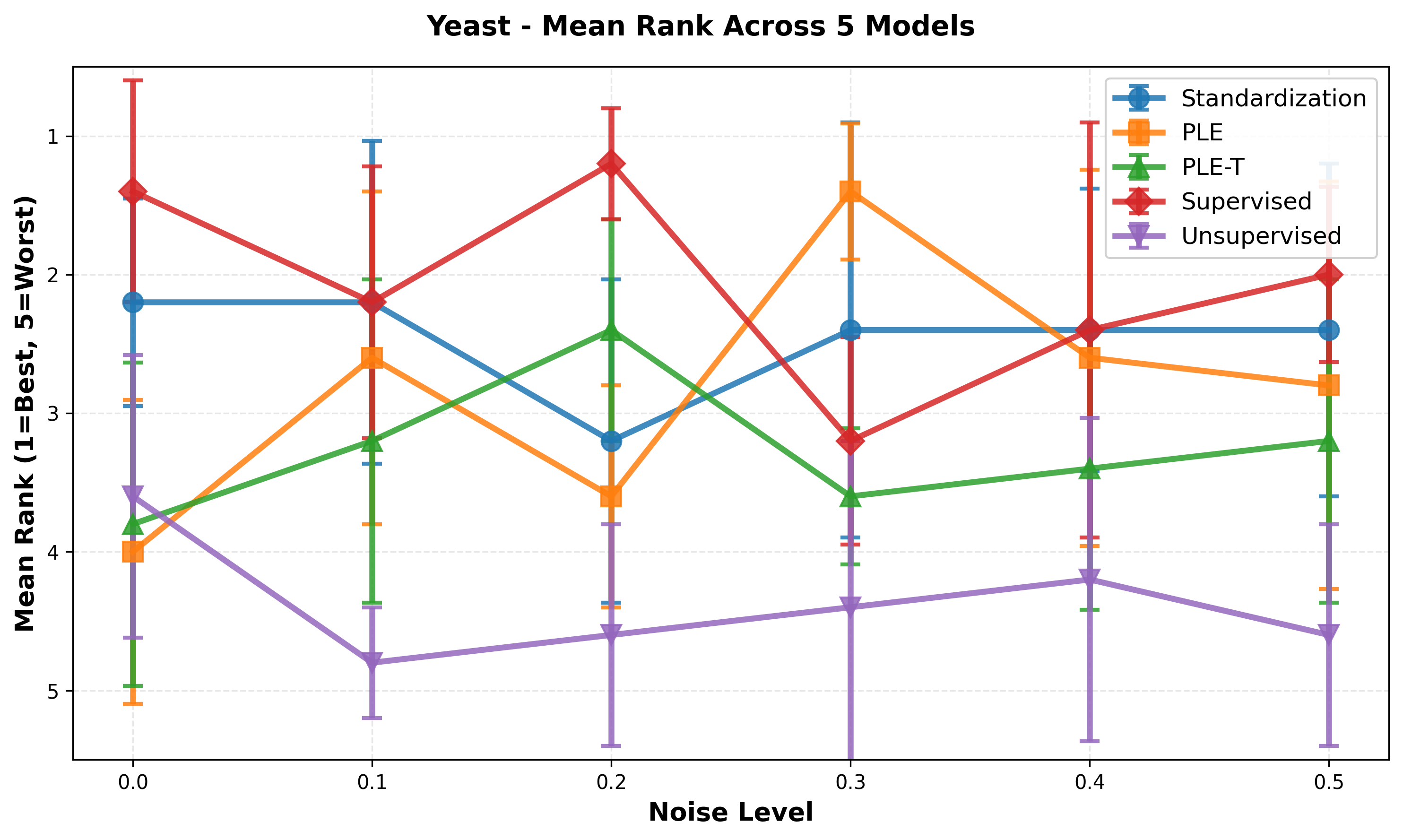}
        \caption{\texttt{yeast} (Classification)}
        \label{fig:n4}
    \end{subfigure}
    \caption{\textbf{Detailed Mean Rank Trajectories (2/2).} Mean rank across 5 models for IBM-Employee and Yeast datasets under varying noise levels.}
    \label{fig:noise_rank_part2}
\end{figure}

Figure~\ref{fig:noise_rank} shows the aggregated mean rank as a function of noise level; Figures~\ref{fig:n1}--\ref{fig:n4} show per-dataset ranks.

\textbf{Supervised Stretch does not collapse under noise.} The mean rank of Supervised Stretch (red) stays in a competitive range over the full noise sweep and does not show the gradual degradation seen for some baselines. At the highest noise level we test ($\eta = 0.5$), it actually has the best mean rank among the five methods. This is consistent with the role of target information in the transformation: bin widths are computed from out-of-fold smoothed estimates of $f(x)$, which average out a substantial fraction of label noise before it ever affects the transformation.

\textbf{Unsupervised Stretch is a stable target-free option.} Standardization (blue) is a flat reference, as expected. Unsupervised Stretch (purple) tracks the same trend across all noise levels and at $\eta = 0.5$ ranks third, slightly behind PLE. Since it uses only $O(1)$ memory per feature instead of PLE's $O(T)$, it remains an efficient choice when one cannot trust the targets enough to use Supervised Stretch.

% \update{\subsection{Results and Discussion}
% Figures~\ref{fig:n1},~\ref{fig:n2} and~\ref{fig:n3} illustrate the performance trajectories as noise intensity increases. We report the mean accuracy and add error bars. The empirical results challenge the conventional intuition that supervised feature transformations are inherently fragile to label noise:}

% \begin{itemize}[leftmargin=1.5em, topsep=0pt, itemsep=0pt]
%     \item \update{\textbf{Supervised Stretch Exhibits Superior Robustness}: Contrary to a potential concern that using target information would lead to overfitting noise, all figures show that Supervised Stretch (red line) demonstrates remarkable stability. In two out of the three datasets (Figures~\ref{fig:n1} and ~\ref{fig:n3}), it remains the superior method at the most extreme noise level (50\%)}. 
%     \item \update{\textbf{Unsupervised Stretch Performance Varies}: The performance of Unsupervised Stretch (purple line) is dataset-dependent. In one case (Figure~\ref{fig:n2}), it eventually surpasses other methods at the highest noise level, serving as a useful fallback when the target signal is virtually destroyed. However, in other cases (e.g., Figure~\ref{fig:n3}), its performance drops significantly alongside other baselines, indicating that simply flattening the feature distribution is not a guaranteed solution for noise resilience.}
% \end{itemize}

%%%%%%%%%%%%%%%%%%%%%%%%%%%%%%%%%%%%%%%%%%%%%%%%%%%%%%%%%%%%

\section{Additional Metrics}
\label{app:additional_metrics}

For completeness, we report aggregated results across seven evaluation metrics: classification (Accuracy, Average Recall, F1, AUC) and regression ($R^2$, MAE, RMSE). Each table summarises one metric across the five models and nine transformations, using the same tie-exclusion protocol as Table~\ref{tab:transformations_performance}: per (dataset, model) pair, if the gap between the best and worst method is at most the median of the per-method seed standard deviations, that pair is treated as a tie and excluded from the per-cell average. The best displayed mean in each model row is shown in bold. The trailing ``$\pm$'' is the standard error of the mean across the remaining datasets within the task category. For MAE and RMSE we caution that cross-dataset aggregation is not particularly meaningful, since these losses depend on each dataset's target scale; we include the aggregated numbers as a coarse summary only.

% Aggregate metric tables (auto-generated)
% One table per metric: 5 models x 9 transformations.
% Tie-exclusion logic matches Table 1 in main.tex.

\begin{table}[t]
\centering
\caption{Aggregate Accuracy per model and transformation (higher is better); a counterpart to the `Avg.~Acc'/`Avg.~$R^2$' rows of Table~\ref{tab:transformations_performance}. Each cell is the mean of the per-(dataset, model) values over the regression or classification benchmarks (with $R^2$ clamped at $0$, matching the main paper). The trailing ``$\pm$'' is the average per-pair standard error of the mean across $15$ seeds, $\sigma_{\rm seed}/\sqrt{15}$, averaged across the contributing datasets; cells whose underlying per-pair seed std exceeds $0.20$ are flagged as ``$>\!.05$''. \textbf{Abbreviations}: Sup.\ (Stretch Supervised), Unsup.\ (Stretch Unsupervised), RS-SC (Robust Scale + Smooth Clip), YJ (Yeo--Johnson).}
\label{tab:agg_accuracy}
\begin{adjustbox}{max width=\textwidth}
\begin{tabular}{c ccccccccc}
\toprule
\multirow{2.5}{*}{\textbf{Model}} & \multicolumn{9}{c}{\textbf{Transformations}} \\[0.0em]
\cmidrule(lr){2-10}
& \makebox[1.2cm][c]{\parbox{1.2cm}{\centering\textbf{Sup.}}} & \makebox[1.2cm][c]{\parbox{1.2cm}{\centering\textbf{Unsup.}}} & \makebox[1.2cm][c]{\parbox{1.2cm}{\centering\textbf{Minmax}}} & \makebox[1.2cm][c]{\parbox{1.2cm}{\centering\textbf{PLE}}} & \makebox[1.2cm][c]{\parbox{1.2cm}{\centering\textbf{PLE-T}}} & \makebox[1.2cm][c]{\parbox{1.2cm}{\centering\textbf{Quantile}}} & \makebox[1.2cm][c]{\parbox{1.2cm}{\centering\textbf{RS-SC}}} & \makebox[1.2cm][c]{\parbox{1.2cm}{\centering\textbf{Standard}}} & \makebox[1.2cm][c]{\parbox{1.2cm}{\centering\textbf{YJ}}} \\
\midrule
    ftt & 0.827{\scriptsize$\pm$.004} & 0.824{\scriptsize$\pm$.003} & 0.805{\scriptsize$\pm$.002} & 0.826{\scriptsize$\pm$.002} & 0.826{\scriptsize$\pm$.002} & \textbf{0.828}{\scriptsize$\pm$.003} & 0.819{\scriptsize$\pm$.003} & 0.816{\scriptsize$\pm$.004} & 0.821{\scriptsize$\pm$.003} \\
    mlp & 0.821{\scriptsize$\pm$.003} & 0.824{\scriptsize$\pm$.002} & 0.815{\scriptsize$\pm$.002} & \textbf{0.828}{\scriptsize$\pm$.002} & 0.822{\scriptsize$\pm$.002} & 0.816{\scriptsize$\pm$.002} & 0.823{\scriptsize$\pm$.002} & 0.819{\scriptsize$\pm$.002} & 0.822{\scriptsize$\pm$.002} \\
    mlp\_plr & 0.826{\scriptsize$\pm$.001} & 0.826{\scriptsize$\pm$.002} & 0.821{\scriptsize$\pm$.002} & 0.821{\scriptsize$\pm$.002} & 0.821{\scriptsize$\pm$.003} & \textbf{0.827}{\scriptsize$\pm$.002} & 0.823{\scriptsize$\pm$.002} & 0.821{\scriptsize$\pm$.002} & 0.821{\scriptsize$\pm$.003} \\
    realmlp & 0.837{\scriptsize$\pm$.002} & 0.838{\scriptsize$\pm$.001} & \textbf{0.839}{\scriptsize$\pm$.002} & 0.833{\scriptsize$\pm$.002} & 0.834{\scriptsize$\pm$.002} & 0.837{\scriptsize$\pm$.002} & 0.838{\scriptsize$\pm$.002} & 0.833{\scriptsize$\pm$.001} & 0.834{\scriptsize$\pm$.002} \\
    resnet & \textbf{0.826}{\scriptsize$\pm$.002} & 0.824{\scriptsize$\pm$.003} & 0.814{\scriptsize$\pm$.003} & 0.824{\scriptsize$\pm$.002} & 0.823{\scriptsize$\pm$.002} & 0.816{\scriptsize$\pm$.002} & 0.825{\scriptsize$\pm$.002} & 0.819{\scriptsize$\pm$.002} & 0.820{\scriptsize$\pm$.002} \\
\bottomrule
\end{tabular}
\end{adjustbox}
\end{table}

\begin{table}[t]
\centering
\caption{Aggregate Avg.\ Recall per model and transformation (higher is better); a counterpart to the `Avg.~Acc'/`Avg.~$R^2$' rows of Table~\ref{tab:transformations_performance}. Each cell is the mean of the per-(dataset, model) values over the regression or classification benchmarks (with $R^2$ clamped at $0$, matching the main paper). The trailing ``$\pm$'' is the average per-pair standard error of the mean across $15$ seeds, $\sigma_{\rm seed}/\sqrt{15}$, averaged across the contributing datasets; cells whose underlying per-pair seed std exceeds $0.20$ are flagged as ``$>\!.05$''. \textbf{Abbreviations}: Sup.\ (Stretch Supervised), Unsup.\ (Stretch Unsupervised), RS-SC (Robust Scale + Smooth Clip), YJ (Yeo--Johnson).}
\label{tab:agg_avg_recall}
\begin{adjustbox}{max width=\textwidth}
\begin{tabular}{c ccccccccc}
\toprule
\multirow{2.5}{*}{\textbf{Model}} & \multicolumn{9}{c}{\textbf{Transformations}} \\[0.0em]
\cmidrule(lr){2-10}
& \makebox[1.2cm][c]{\parbox{1.2cm}{\centering\textbf{Sup.}}} & \makebox[1.2cm][c]{\parbox{1.2cm}{\centering\textbf{Unsup.}}} & \makebox[1.2cm][c]{\parbox{1.2cm}{\centering\textbf{Minmax}}} & \makebox[1.2cm][c]{\parbox{1.2cm}{\centering\textbf{PLE}}} & \makebox[1.2cm][c]{\parbox{1.2cm}{\centering\textbf{PLE-T}}} & \makebox[1.2cm][c]{\parbox{1.2cm}{\centering\textbf{Quantile}}} & \makebox[1.2cm][c]{\parbox{1.2cm}{\centering\textbf{RS-SC}}} & \makebox[1.2cm][c]{\parbox{1.2cm}{\centering\textbf{Standard}}} & \makebox[1.2cm][c]{\parbox{1.2cm}{\centering\textbf{YJ}}} \\
\midrule
    ftt & 0.737{\scriptsize$\pm$.007} & 0.740{\scriptsize$\pm$.004} & 0.711{\scriptsize$\pm$.004} & 0.744{\scriptsize$\pm$.003} & 0.744{\scriptsize$\pm$.004} & \textbf{0.746}{\scriptsize$\pm$.005} & 0.728{\scriptsize$\pm$.005} & 0.728{\scriptsize$\pm$.006} & 0.737{\scriptsize$\pm$.005} \\
    mlp & 0.733{\scriptsize$\pm$.005} & 0.738{\scriptsize$\pm$.004} & 0.723{\scriptsize$\pm$.003} & 0.734{\scriptsize$\pm$\,$>$\,.05} & 0.739{\scriptsize$\pm$.003} & 0.728{\scriptsize$\pm$.004} & \textbf{0.740}{\scriptsize$\pm$.004} & 0.728{\scriptsize$\pm$.004} & 0.733{\scriptsize$\pm$.003} \\
    mlp\_plr & \textbf{0.744}{\scriptsize$\pm$.002} & 0.742{\scriptsize$\pm$.003} & 0.734{\scriptsize$\pm$.003} & 0.733{\scriptsize$\pm$.003} & 0.734{\scriptsize$\pm$.004} & 0.742{\scriptsize$\pm$.005} & 0.740{\scriptsize$\pm$.003} & 0.731{\scriptsize$\pm$.003} & 0.736{\scriptsize$\pm$.004} \\
    realmlp & \textbf{0.762}{\scriptsize$\pm$.003} & 0.760{\scriptsize$\pm$.003} & 0.759{\scriptsize$\pm$.003} & 0.754{\scriptsize$\pm$.003} & 0.757{\scriptsize$\pm$.003} & 0.759{\scriptsize$\pm$.003} & 0.759{\scriptsize$\pm$.003} & 0.747{\scriptsize$\pm$.003} & 0.751{\scriptsize$\pm$.003} \\
    resnet & \textbf{0.746}{\scriptsize$\pm$.003} & 0.740{\scriptsize$\pm$.005} & 0.728{\scriptsize$\pm$.005} & 0.738{\scriptsize$\pm$.003} & 0.740{\scriptsize$\pm$.003} & 0.729{\scriptsize$\pm$.004} & 0.744{\scriptsize$\pm$.004} & 0.732{\scriptsize$\pm$.003} & 0.734{\scriptsize$\pm$.004} \\
\bottomrule
\end{tabular}
\end{adjustbox}
\end{table}

\begin{table}[t]
\centering
\caption{Aggregate F1 per model and transformation (higher is better); a counterpart to the `Avg.~Acc'/`Avg.~$R^2$' rows of Table~\ref{tab:transformations_performance}. Each cell is the mean of the per-(dataset, model) values over the regression or classification benchmarks (with $R^2$ clamped at $0$, matching the main paper). The trailing ``$\pm$'' is the average per-pair standard error of the mean across $15$ seeds, $\sigma_{\rm seed}/\sqrt{15}$, averaged across the contributing datasets; cells whose underlying per-pair seed std exceeds $0.20$ are flagged as ``$>\!.05$''. \textbf{Abbreviations}: Sup.\ (Stretch Supervised), Unsup.\ (Stretch Unsupervised), RS-SC (Robust Scale + Smooth Clip), YJ (Yeo--Johnson).}
\label{tab:agg_f1}
\begin{adjustbox}{max width=\textwidth}
\begin{tabular}{c ccccccccc}
\toprule
\multirow{2.5}{*}{\textbf{Model}} & \multicolumn{9}{c}{\textbf{Transformations}} \\[0.0em]
\cmidrule(lr){2-10}
& \makebox[1.2cm][c]{\parbox{1.2cm}{\centering\textbf{Sup.}}} & \makebox[1.2cm][c]{\parbox{1.2cm}{\centering\textbf{Unsup.}}} & \makebox[1.2cm][c]{\parbox{1.2cm}{\centering\textbf{Minmax}}} & \makebox[1.2cm][c]{\parbox{1.2cm}{\centering\textbf{PLE}}} & \makebox[1.2cm][c]{\parbox{1.2cm}{\centering\textbf{PLE-T}}} & \makebox[1.2cm][c]{\parbox{1.2cm}{\centering\textbf{Quantile}}} & \makebox[1.2cm][c]{\parbox{1.2cm}{\centering\textbf{RS-SC}}} & \makebox[1.2cm][c]{\parbox{1.2cm}{\centering\textbf{Standard}}} & \makebox[1.2cm][c]{\parbox{1.2cm}{\centering\textbf{YJ}}} \\
\midrule
    ftt & 0.680{\scriptsize$\pm$\,$>$\,.05} & 0.681{\scriptsize$\pm$\,$>$\,.05} & 0.656{\scriptsize$\pm$.005} & 0.691{\scriptsize$\pm$.005} & 0.689{\scriptsize$\pm$.006} & \textbf{0.693}{\scriptsize$\pm$.006} & 0.672{\scriptsize$\pm$.006} & 0.665{\scriptsize$\pm$\,$>$\,.05} & 0.683{\scriptsize$\pm$\,$>$\,.05} \\
    mlp & 0.670{\scriptsize$\pm$\,$>$\,.05} & 0.682{\scriptsize$\pm$.005} & 0.662{\scriptsize$\pm$.003} & 0.682{\scriptsize$\pm$\,$>$\,.05} & \textbf{0.686}{\scriptsize$\pm$.005} & 0.673{\scriptsize$\pm$.005} & \textbf{0.686}{\scriptsize$\pm$.005} & 0.670{\scriptsize$\pm$.005} & 0.674{\scriptsize$\pm$.004} \\
    mlp\_plr & \textbf{0.693}{\scriptsize$\pm$.003} & 0.692{\scriptsize$\pm$.004} & 0.676{\scriptsize$\pm$.004} & 0.684{\scriptsize$\pm$.004} & 0.680{\scriptsize$\pm$.005} & 0.688{\scriptsize$\pm$.005} & 0.690{\scriptsize$\pm$.004} & 0.675{\scriptsize$\pm$.004} & 0.683{\scriptsize$\pm$.005} \\
    realmlp & \textbf{0.713}{\scriptsize$\pm$.004} & 0.709{\scriptsize$\pm$.004} & 0.709{\scriptsize$\pm$.004} & 0.704{\scriptsize$\pm$.003} & 0.707{\scriptsize$\pm$.003} & 0.709{\scriptsize$\pm$.004} & 0.709{\scriptsize$\pm$.003} & 0.691{\scriptsize$\pm$.004} & 0.699{\scriptsize$\pm$.003} \\
    resnet & 0.692{\scriptsize$\pm$.004} & 0.679{\scriptsize$\pm$\,$>$\,.05} & 0.670{\scriptsize$\pm$.006} & 0.686{\scriptsize$\pm$.004} & 0.687{\scriptsize$\pm$.004} & 0.670{\scriptsize$\pm$.006} & \textbf{0.693}{\scriptsize$\pm$.006} & 0.670{\scriptsize$\pm$.004} & 0.677{\scriptsize$\pm$.006} \\
\bottomrule
\end{tabular}
\end{adjustbox}
\end{table}

\begin{table}[t]
\centering
\caption{Aggregate AUC per model and transformation (higher is better); a counterpart to the `Avg.~Acc'/`Avg.~$R^2$' rows of Table~\ref{tab:transformations_performance}. Each cell is the mean of the per-(dataset, model) values over the regression or classification benchmarks (with $R^2$ clamped at $0$, matching the main paper). The trailing ``$\pm$'' is the average per-pair standard error of the mean across $15$ seeds, $\sigma_{\rm seed}/\sqrt{15}$, averaged across the contributing datasets; cells whose underlying per-pair seed std exceeds $0.20$ are flagged as ``$>\!.05$''. \textbf{Abbreviations}: Sup.\ (Stretch Supervised), Unsup.\ (Stretch Unsupervised), RS-SC (Robust Scale + Smooth Clip), YJ (Yeo--Johnson).}
\label{tab:agg_auc}
\begin{adjustbox}{max width=\textwidth}
\begin{tabular}{c ccccccccc}
\toprule
\multirow{2.5}{*}{\textbf{Model}} & \multicolumn{9}{c}{\textbf{Transformations}} \\[0.0em]
\cmidrule(lr){2-10}
& \makebox[1.2cm][c]{\parbox{1.2cm}{\centering\textbf{Sup.}}} & \makebox[1.2cm][c]{\parbox{1.2cm}{\centering\textbf{Unsup.}}} & \makebox[1.2cm][c]{\parbox{1.2cm}{\centering\textbf{Minmax}}} & \makebox[1.2cm][c]{\parbox{1.2cm}{\centering\textbf{PLE}}} & \makebox[1.2cm][c]{\parbox{1.2cm}{\centering\textbf{PLE-T}}} & \makebox[1.2cm][c]{\parbox{1.2cm}{\centering\textbf{Quantile}}} & \makebox[1.2cm][c]{\parbox{1.2cm}{\centering\textbf{RS-SC}}} & \makebox[1.2cm][c]{\parbox{1.2cm}{\centering\textbf{Standard}}} & \makebox[1.2cm][c]{\parbox{1.2cm}{\centering\textbf{YJ}}} \\
\midrule
    ftt & 0.868{\scriptsize$\pm$.007} & 0.866{\scriptsize$\pm$.004} & 0.834{\scriptsize$\pm$.005} & \textbf{0.872}{\scriptsize$\pm$.002} & 0.869{\scriptsize$\pm$.004} & \textbf{0.872}{\scriptsize$\pm$.004} & 0.862{\scriptsize$\pm$.004} & 0.859{\scriptsize$\pm$.005} & 0.862{\scriptsize$\pm$.005} \\
    mlp & 0.861{\scriptsize$\pm$.004} & \textbf{0.871}{\scriptsize$\pm$.002} & 0.855{\scriptsize$\pm$.002} & 0.866{\scriptsize$\pm$.003} & 0.861{\scriptsize$\pm$.004} & 0.868{\scriptsize$\pm$.002} & 0.866{\scriptsize$\pm$.003} & 0.865{\scriptsize$\pm$.002} & 0.867{\scriptsize$\pm$.002} \\
    mlp\_plr & 0.863{\scriptsize$\pm$.002} & 0.863{\scriptsize$\pm$.003} & 0.866{\scriptsize$\pm$.002} & 0.856{\scriptsize$\pm$.002} & 0.860{\scriptsize$\pm$.003} & \textbf{0.867}{\scriptsize$\pm$.003} & 0.863{\scriptsize$\pm$.003} & 0.859{\scriptsize$\pm$.003} & 0.862{\scriptsize$\pm$.003} \\
    realmlp & 0.867{\scriptsize$\pm$.003} & 0.871{\scriptsize$\pm$.003} & \textbf{0.875}{\scriptsize$\pm$.002} & 0.863{\scriptsize$\pm$.003} & 0.872{\scriptsize$\pm$.003} & 0.873{\scriptsize$\pm$.003} & 0.871{\scriptsize$\pm$.003} & 0.873{\scriptsize$\pm$.003} & 0.872{\scriptsize$\pm$.003} \\
    resnet & \textbf{0.872}{\scriptsize$\pm$.003} & 0.863{\scriptsize$\pm$\,$>$\,.05} & 0.858{\scriptsize$\pm$.005} & 0.869{\scriptsize$\pm$.002} & 0.864{\scriptsize$\pm$.003} & \textbf{0.872}{\scriptsize$\pm$.003} & \textbf{0.872}{\scriptsize$\pm$.005} & 0.869{\scriptsize$\pm$.003} & \textbf{0.872}{\scriptsize$\pm$.003} \\
\bottomrule
\end{tabular}
\end{adjustbox}
\end{table}

\begin{table}[t]
\centering
\caption{Aggregate $R^2$ per model and transformation (higher is better); a counterpart to the `Avg.~Acc'/`Avg.~$R^2$' rows of Table~\ref{tab:transformations_performance}. Each cell is the mean of the per-(dataset, model) values over the regression or classification benchmarks (with $R^2$ clamped at $0$, matching the main paper). The trailing ``$\pm$'' is the average per-pair standard error of the mean across $15$ seeds, $\sigma_{\rm seed}/\sqrt{15}$, averaged across the contributing datasets; cells whose underlying per-pair seed std exceeds $0.20$ are flagged as ``$>\!.05$''. \textbf{Abbreviations}: Sup.\ (Stretch Supervised), Unsup.\ (Stretch Unsupervised), RS-SC (Robust Scale + Smooth Clip), YJ (Yeo--Johnson).}
\label{tab:agg_r2}
\begin{adjustbox}{max width=\textwidth}
\begin{tabular}{c ccccccccc}
\toprule
\multirow{2.5}{*}{\textbf{Model}} & \multicolumn{9}{c}{\textbf{Transformations}} \\[0.0em]
\cmidrule(lr){2-10}
& \makebox[1.2cm][c]{\parbox{1.2cm}{\centering\textbf{Sup.}}} & \makebox[1.2cm][c]{\parbox{1.2cm}{\centering\textbf{Unsup.}}} & \makebox[1.2cm][c]{\parbox{1.2cm}{\centering\textbf{Minmax}}} & \makebox[1.2cm][c]{\parbox{1.2cm}{\centering\textbf{PLE}}} & \makebox[1.2cm][c]{\parbox{1.2cm}{\centering\textbf{PLE-T}}} & \makebox[1.2cm][c]{\parbox{1.2cm}{\centering\textbf{Quantile}}} & \makebox[1.2cm][c]{\parbox{1.2cm}{\centering\textbf{RS-SC}}} & \makebox[1.2cm][c]{\parbox{1.2cm}{\centering\textbf{Standard}}} & \makebox[1.2cm][c]{\parbox{1.2cm}{\centering\textbf{YJ}}} \\
\midrule
    ftt & \textbf{0.691}{\scriptsize$\pm$.002} & 0.681{\scriptsize$\pm$.002} & 0.666{\scriptsize$\pm$.003} & 0.680{\scriptsize$\pm$.006} & 0.667{\scriptsize$\pm$.003} & 0.681{\scriptsize$\pm$.002} & 0.673{\scriptsize$\pm$.003} & 0.681{\scriptsize$\pm$.001} & 0.667{\scriptsize$\pm$.001} \\
    mlp & \textbf{0.630}{\scriptsize$\pm$.005} & 0.606{\scriptsize$\pm$.007} & 0.628{\scriptsize$\pm$.008} & 0.608{\scriptsize$\pm$\,$>$\,.05} & 0.588{\scriptsize$\pm$\,$>$\,.05} & 0.619{\scriptsize$\pm$.005} & 0.619{\scriptsize$\pm$.005} & 0.591{\scriptsize$\pm$\,$>$\,.05} & 0.545{\scriptsize$\pm$\,$>$\,.05} \\
    mlp\_plr & 0.698{\scriptsize$\pm$.004} & 0.696{\scriptsize$\pm$.004} & 0.659{\scriptsize$\pm$.004} & 0.686{\scriptsize$\pm$.002} & \textbf{0.702}{\scriptsize$\pm$.002} & 0.669{\scriptsize$\pm$.005} & 0.681{\scriptsize$\pm$.004} & 0.675{\scriptsize$\pm$.003} & 0.649{\scriptsize$\pm$.005} \\
    realmlp & \textbf{0.736}{\scriptsize$\pm$.004} & 0.685{\scriptsize$\pm$.004} & 0.679{\scriptsize$\pm$.003} & \textbf{0.736}{\scriptsize$\pm$.003} & 0.723{\scriptsize$\pm$.003} & 0.707{\scriptsize$\pm$.005} & 0.689{\scriptsize$\pm$.004} & 0.686{\scriptsize$\pm$.002} & 0.683{\scriptsize$\pm$\,$>$\,.05} \\
    resnet & 0.651{\scriptsize$\pm$.003} & 0.631{\scriptsize$\pm$.006} & \textbf{0.655}{\scriptsize$\pm$.002} & 0.639{\scriptsize$\pm$\,$>$\,.05} & 0.631{\scriptsize$\pm$.005} & 0.625{\scriptsize$\pm$\,$>$\,.05} & 0.640{\scriptsize$\pm$.002} & 0.641{\scriptsize$\pm$.003} & 0.542{\scriptsize$\pm$\,$>$\,.05} \\
\bottomrule
\end{tabular}
\end{adjustbox}
\end{table}

\begin{table}[t]
\centering
\caption{Aggregate MAE per model and transformation (lower is better); a counterpart to the `Avg.~Acc'/`Avg.~$R^2$' rows of Table~\ref{tab:transformations_performance}. Each cell is the mean of the per-(dataset, model) values over the regression benchmarks. Note: MAE and RMSE are tied to each dataset's target scale, so the cross-dataset average is only a coarse summary and is not directly comparable across rows; we include it for completeness only and omit a $\pm$ uncertainty. \textbf{Abbreviations}: Sup.\ (Stretch Supervised), Unsup.\ (Stretch Unsupervised), RS-SC (Robust Scale + Smooth Clip), YJ (Yeo--Johnson).}
\label{tab:agg_mae}
\begin{adjustbox}{max width=\textwidth}
\begin{tabular}{c ccccccccc}
\toprule
\multirow{2.5}{*}{\textbf{Model}} & \multicolumn{9}{c}{\textbf{Transformations}} \\[0.0em]
\cmidrule(lr){2-10}
& \makebox[1.2cm][c]{\parbox{1.2cm}{\centering\textbf{Sup.}}} & \makebox[1.2cm][c]{\parbox{1.2cm}{\centering\textbf{Unsup.}}} & \makebox[1.2cm][c]{\parbox{1.2cm}{\centering\textbf{Minmax}}} & \makebox[1.2cm][c]{\parbox{1.2cm}{\centering\textbf{PLE}}} & \makebox[1.2cm][c]{\parbox{1.2cm}{\centering\textbf{PLE-T}}} & \makebox[1.2cm][c]{\parbox{1.2cm}{\centering\textbf{Quantile}}} & \makebox[1.2cm][c]{\parbox{1.2cm}{\centering\textbf{RS-SC}}} & \makebox[1.2cm][c]{\parbox{1.2cm}{\centering\textbf{Standard}}} & \makebox[1.2cm][c]{\parbox{1.2cm}{\centering\textbf{YJ}}} \\
\midrule
    ftt & 30.84 & 30.80 & 31.27 & 30.41 & \textbf{30.27} & 31.10 & 30.90 & 30.73 & 31.62 \\
    mlp & 36.42 & 37.19 & 36.91 & 38.50 & 37.25 & 36.97 & \textbf{36.03} & 37.88 & 36.29 \\
    mlp\_plr & 33.06 & 32.96 & \textbf{32.58} & 34.91 & 33.96 & 34.07 & 33.16 & 33.59 & 33.47 \\
    realmlp & 30.10 & 30.31 & 30.79 & \textbf{30.09} & 30.33 & 30.28 & 30.25 & 30.33 & 30.79 \\
    resnet & 35.16 & 35.32 & 35.67 & 34.71 & 34.56 & 36.34 & \textbf{34.26} & 35.47 & 37.42 \\
\bottomrule
\end{tabular}
\end{adjustbox}
\end{table}

\begin{table}[t]
\centering
\caption{Aggregate RMSE per model and transformation (lower is better); a counterpart to the `Avg.~Acc'/`Avg.~$R^2$' rows of Table~\ref{tab:transformations_performance}. Each cell is the mean of the per-(dataset, model) values over the regression benchmarks. Note: MAE and RMSE are tied to each dataset's target scale, so the cross-dataset average is only a coarse summary and is not directly comparable across rows; we include it for completeness only and omit a $\pm$ uncertainty. \textbf{Abbreviations}: Sup.\ (Stretch Supervised), Unsup.\ (Stretch Unsupervised), RS-SC (Robust Scale + Smooth Clip), YJ (Yeo--Johnson).}
\label{tab:agg_rmse}
\begin{adjustbox}{max width=\textwidth}
\begin{tabular}{c ccccccccc}
\toprule
\multirow{2.5}{*}{\textbf{Model}} & \multicolumn{9}{c}{\textbf{Transformations}} \\[0.0em]
\cmidrule(lr){2-10}
& \makebox[1.2cm][c]{\parbox{1.2cm}{\centering\textbf{Sup.}}} & \makebox[1.2cm][c]{\parbox{1.2cm}{\centering\textbf{Unsup.}}} & \makebox[1.2cm][c]{\parbox{1.2cm}{\centering\textbf{Minmax}}} & \makebox[1.2cm][c]{\parbox{1.2cm}{\centering\textbf{PLE}}} & \makebox[1.2cm][c]{\parbox{1.2cm}{\centering\textbf{PLE-T}}} & \makebox[1.2cm][c]{\parbox{1.2cm}{\centering\textbf{Quantile}}} & \makebox[1.2cm][c]{\parbox{1.2cm}{\centering\textbf{RS-SC}}} & \makebox[1.2cm][c]{\parbox{1.2cm}{\centering\textbf{Standard}}} & \makebox[1.2cm][c]{\parbox{1.2cm}{\centering\textbf{YJ}}} \\
\midrule
    ftt & 57.11 & 57.28 & 57.97 & 56.97 & \textbf{56.45} & 57.80 & 57.40 & 57.13 & 59.47 \\
    mlp & 67.4 & 128.2 & 74.7 & 123.9 & 100.6 & 69.5 & \textbf{66.1} & 204.7 & 114.6 \\
    mlp\_plr & 61.25 & 60.00 & \textbf{59.69} & 63.26 & 62.58 & 62.27 & 61.70 & 61.19 & 62.85 \\
    realmlp & \textbf{55.34} & 56.84 & 56.41 & 56.63 & 56.93 & 56.43 & 55.87 & 56.85 & 57.58 \\
    resnet & \textbf{62.1} & 94.9 & 74.8 & 113.0 & 113.2 & 65.5 & \textbf{62.1} & 79.2 & 177.2 \\
\bottomrule
\end{tabular}
\end{adjustbox}
\end{table}

% \newpage
% \clearpage
% \input{checklist.tex}

\end{document}